\documentclass{article} 
\usepackage{nips15submit_e,times}
\usepackage{hyperref}
\usepackage{url}
\usepackage{epsfig}
\usepackage{epstopdf}
\usepackage{graphicx}
\usepackage{amsmath}
\usepackage{amssymb}

\usepackage{algorithm}
\usepackage{algorithmic}
\usepackage{caption}
\usepackage{subcaption}
\usepackage{bm}

\usepackage[numbers,sort]{natbib}

\title{Fast ADMM Algorithm for Distributed Optimization with Adaptive Penalty}

\author{
Changkyu Song \And Sejong Yoon \And Vladimir Pavlovic \\
Department of Computer Science\\
Rutgers, The State University of New Jersey\\
\texttt{\{cs1080, sjyoon, vladimir\}@cs.rutgers.edu}
}

\nipsfinalcopy 

\begin{document}

\maketitle

\begin{abstract}
We propose new methods to speed up convergence of the Alternating Direction Method of Multipliers (ADMM), a common optimization tool in the context of large scale and distributed learning. The proposed method accelerates the speed of convergence by automatically deciding the constraint penalty needed for parameter consensus in each iteration. In addition, we also propose an extension of the method that adaptively determines the maximum number of iterations to update the penalty. We show that this approach effectively leads to an adaptive, dynamic network topology underlying the distributed optimization. The utility of the new penalty update schemes is demonstrated on both synthetic and real data, including a computer vision application of distributed structure from motion.
\end{abstract}

\section{Introduction}


The need for algorithms and methods that can handle large data in a distributed setting has grown significantly in recent years. Specifically, such settings may arise in two prototypical scenarios: (a) induced distributed data: distribute and parallelize computationally demanding optimization tasks to connected computational nodes using a data distributed model and (b) intrinsically distributed data: data is collected across a connected network of sensors (e.g., mobile devices, camera networks), where some or all of the computation can be performed in individual sensor nodes without requiring centralized data pooling. Several distributed learning approaches have been proposed to meet these needs. In particular, the alternating direction method of multiplier (ADMM)~\cite{boyd2010} is an optimization technique that has been very often used in computer vision and machine learning to handle model estimation and learning in either of the two large data settings~\cite{risheng2012, liansheng2012, ehsan2013, zinan2013,  chunyu2014, lai2014, boussaid2014, miksik2014}.

In the distributed optimization setting, the distributed nodes process data locally by solving small optimization problems and aggregate the result by exchanging the (possibly compressed) local solutions (e.g., local model parameter estimates) to arrive at a consensus global result. However, the nature of distributed learning models, particularly in the fully distributed setting where no network topology is presumed, inherently requires repetitive communications between the device nodes. Therefore, it is desirable to reduce the amount of information exchanged and simultaneously improve computational efficiency through faster convergence of such distributed algorithms. 

To this end, the contributions of this paper are three fold. 
\begin{itemize}
\item We propose two variants of ADMM for the consensus-based distributed learning faster than the standard ADMM. Our method extends an acceleration approach for ADMM~\cite{he2000} by an efficient variable penalty parameter update strategy. This strategy results in improved convergence properties of ADMM and also works in a fully distributed fashion.
\item We extend our proposed method to automatically determine the maximum number of iterations allocated to successive updates by employing a budget magement scheme.  This strategy results in adaptive parameter tuning for ADMM, removing the need for arbitrary parameter settings, and effectively induces a varying network communication topology.
\item We apply the proposed method to a prototypical vision and learning problem, the distributed PPCA for structure-from-motion, and demonstrate its empirical utility over the traditional ADMM. 
\end{itemize}

\section{Problem Description and Related Works}

\begin{figure*}[t]
	\centering
	\begin{subfigure}[h]{0.28\textwidth}
		\includegraphics[trim=1cm 1cm 0cm 4cm, width=1\textwidth]{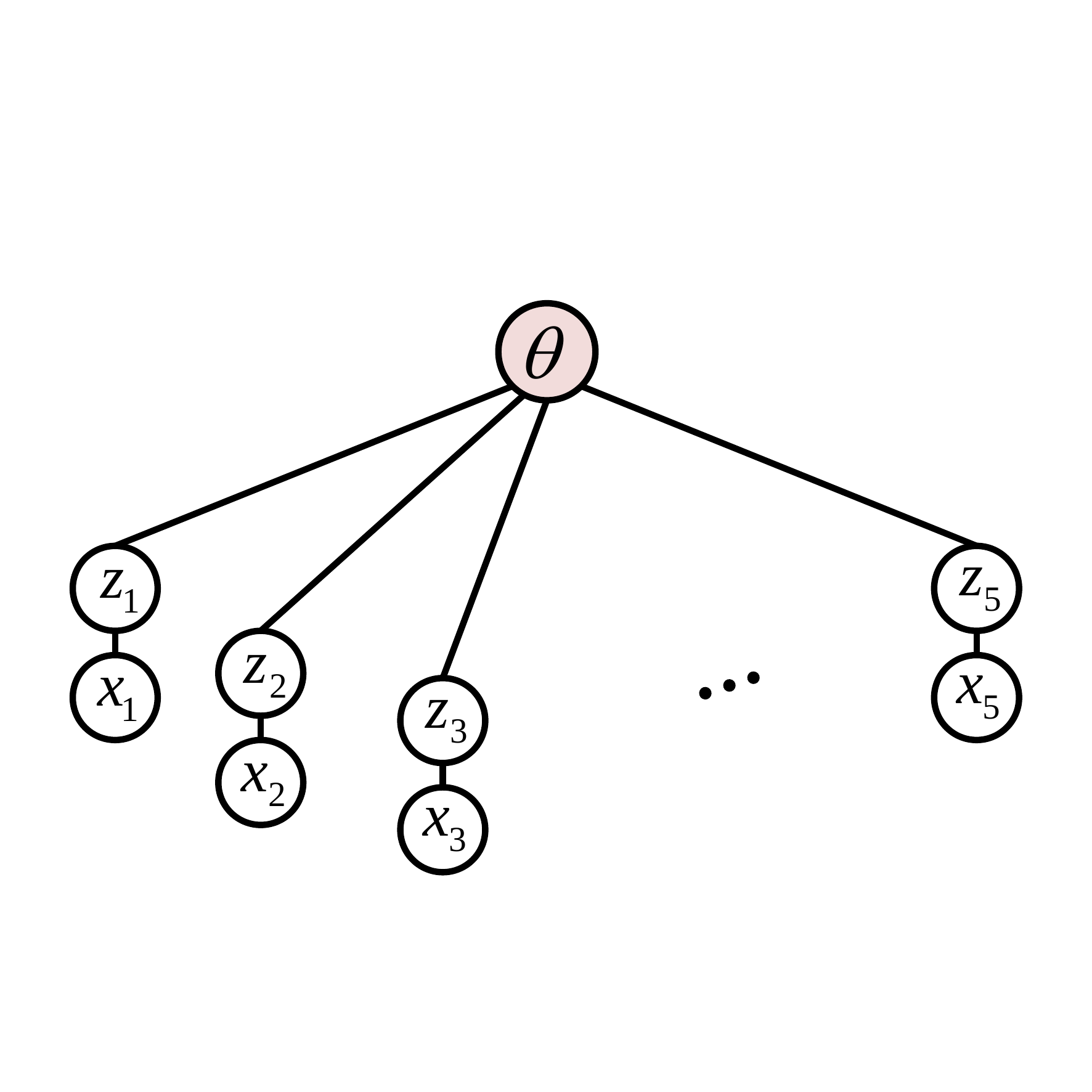}
		\caption{Centralized}
		\label{fig1:cent}
	\end{subfigure}%
	\qquad 
	\begin{subfigure}[h]{0.28\textwidth}
		\includegraphics[trim=1cm 3cm 0cm 2cm, width=1\textwidth]{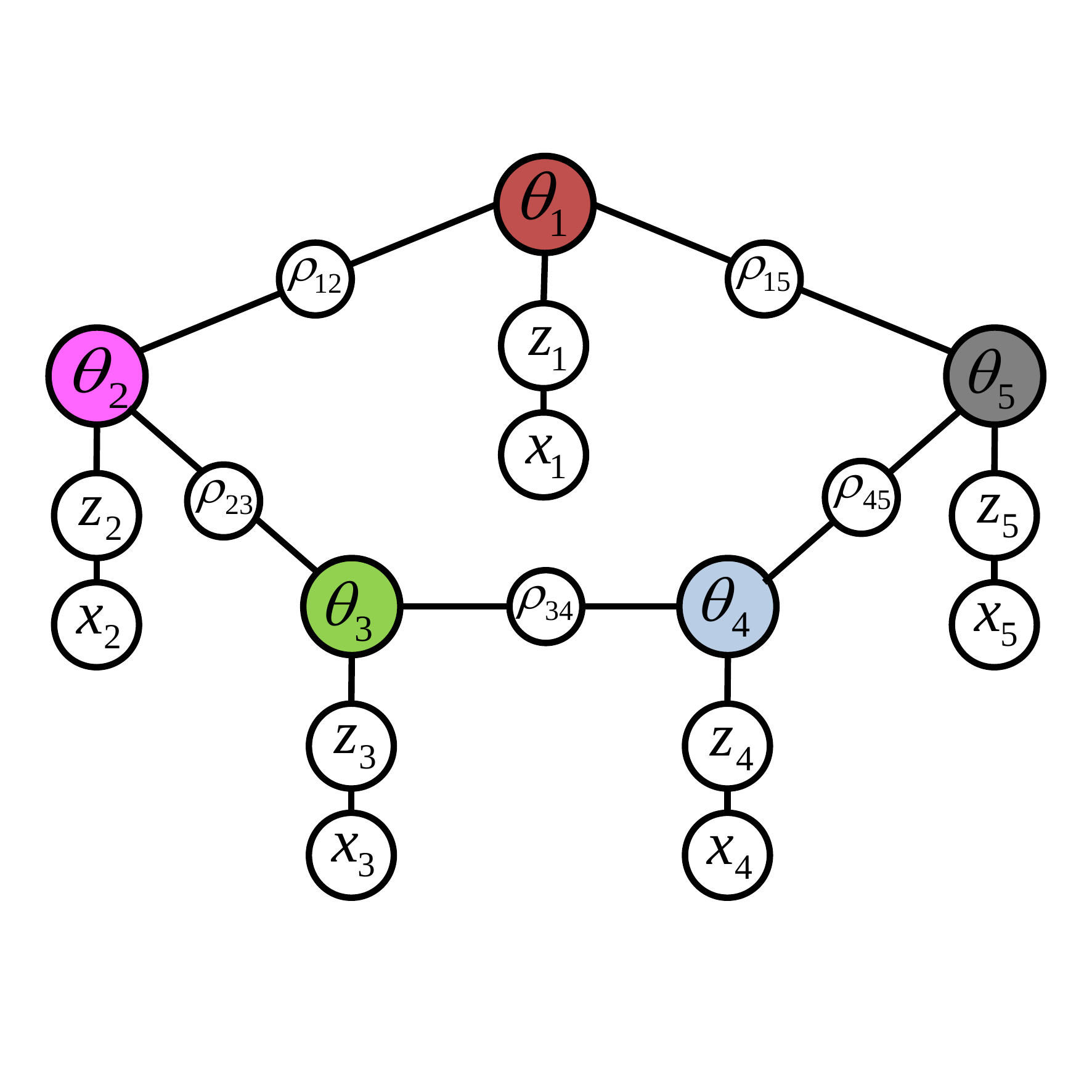}
		\caption{Distributed}
		\label{fig1:dist}
	\end{subfigure}
	\qquad 
	\begin{subfigure}[h]{0.28\textwidth}
		\includegraphics[trim=1cm 3cm 0cm 2cm, width=1\textwidth]{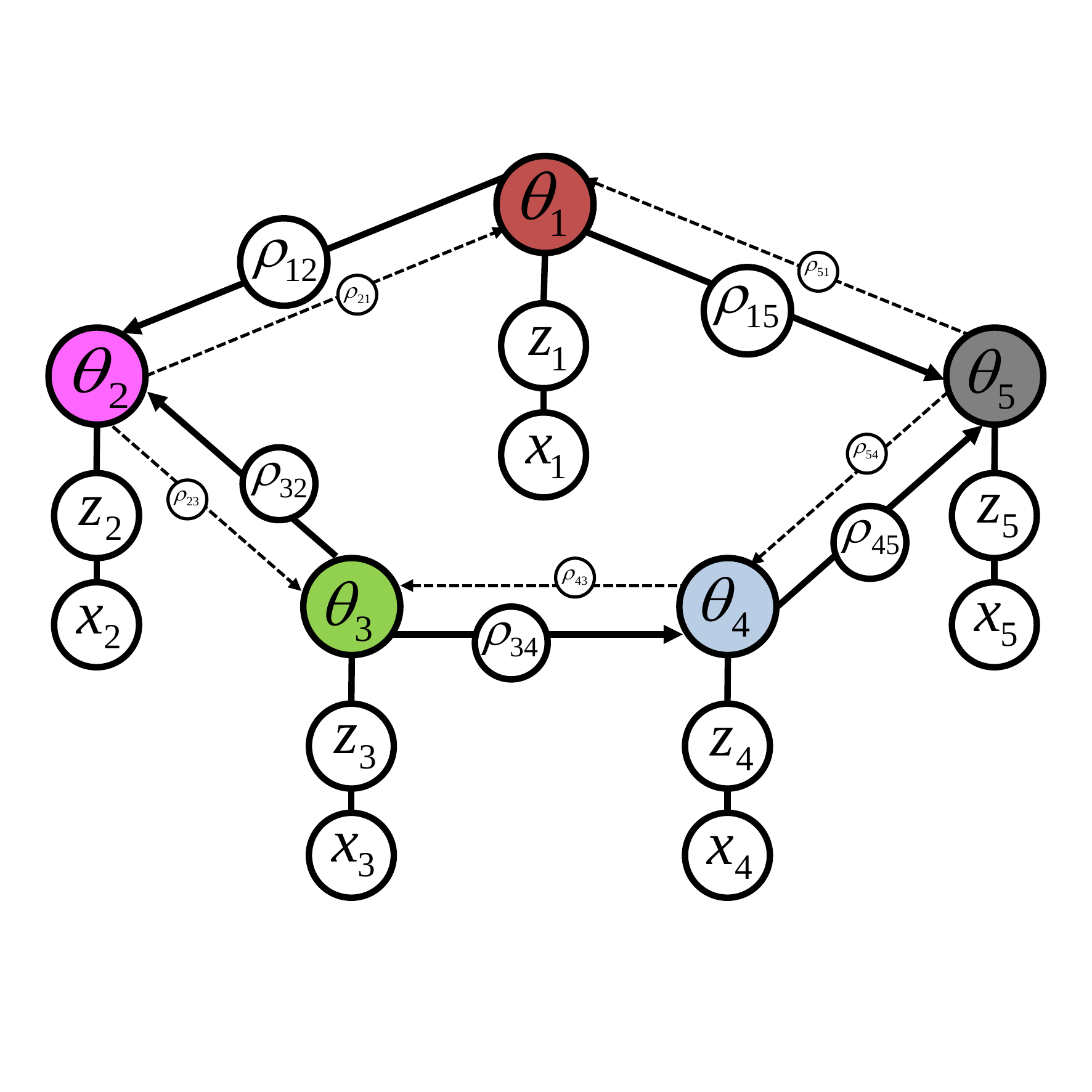}
		\caption{Proposed}
		\label{fig1:proposed}
	\end{subfigure}
	\caption{Centralized, distributed, and the proposed learning model in a ring network. The bigger size of $\rho_{ij}$ means that corresponding constraint is more penalized. Solid edges denote currently strongly influencing edges and dotted edges indicate the edges with less influence.}
	\label{fig:CPL_DPL_DPLANT}
\end{figure*}

The problem we consider in this paper can be formulated as a consensus-based optimization problem~\cite{bertsekas1989}. A general consensus-based optimization problem can be written as
\begin{align}
\label{eq:obj_cent}
\arg\min_{\theta_{i}} &\quad \sum_{i=1}^{J} f_{i}(\theta_{i}),\quad s.t. \quad \theta_{i} = \theta_{j}, \forall i \neq j
\end{align}
where we want to find the set of optimal parameters $\theta_{i}, i = 1..J$ that minimizes the sum of convex objective functions $f_{i}(\theta_{i})$, where $J$ denotes the total number of the functions. This problem is typically a reformulation of a centralized  optimization task $\arg \min f(\theta)$ with a decomposable objective $ f(\theta) =\sum_{i=1}^J f_i(\theta)$.   Given the consensus formulation, the original problem can be solved by decomposing the problem into $J$ subproblems so that $J$ processors can cooperate to solve the overall problem by changing the equality constraint to $\theta_{i} = \bar{\theta}$ where $\bar{\theta}$ denotes a globally shared parameter. The optimization can be approached efficiently by exploiting the alternating direction method of multiplier (ADMM)~\cite{boyd2010}. 

The above consensus formulation is particularly suitable for  many optimization problems that appear in computer vision.  For instance, since $f_{i}(\theta_{i})$ can be any convex function, we can also consider a probabilistic model with the joint negative log likelihood $f_{i}(\theta_{i}) = -\log p(x_{i}, z_{i} | \theta_{i})$ between the observation $x_{i}$ and the corresponding latent variable $z_{i}$. Assuming $(x_{i}, z_{i})$ are independent and identically distributed, finding the maximum likelihood estimate of the shared paramter $\bar{\theta}$ can then be formulated as the optimization problem we described above for many exponential family parametric densities. Moreover, the function need not be a likelihood, but can also be a typical decomposable and regularized loss that occurs in many vision problems such as denoising or dictionary learning.

It is often very convenient to consider the above consensus optimization problem from the perspective of optimization on graphs.  For instance, the centralized i.i.d. Maximum Likelihood learning can be viewed as the optimization on the graph in Fig.~\ref{fig1:cent}. Edges in this graph depict functional (in)dependencies among variables, commonly found in representations such as Markov Random Fields \cite{miksik2014} or Factor Graphs \cite{bishop2006}.  In this context, to fully decompose $f(\cdot)$ and eliminate the need for a processing center completely, one can introduce auxiliary variables $\rho_{ij}$ on every edge to break the dependency between $\theta_{i}$ and $\theta_{j}$~\cite{forero2011, yoon2012} as shown in Fig.~\ref{fig1:dist}. This generalizes to arbitrary graphs, where the connectivity structure may be implied by node placement or communication constraints (camera networks), imaging constraints (pixel neighborhoods in images or frames in a video sequence), or other contextual constraints (loss and regularization structure). 

In general, given a connected graph $\mathcal{G} = (\mathcal{V}, \mathcal{E})$ with the nodes $i, j \in \mathcal{V}$ and the edges $e_{ij} = (i,j) \in \mathcal{E}$, the consensus optimization problem becomes 
\begin{align}
\min \sum_{i \in \mathcal{V}} f_{i}(\theta_{i}), \quad s.t. \quad \theta_{i} = \rho_{ij}, \rho_{ij} = \theta_{j}, j \in \mathcal{B}_{i}
\end{align}
Solving that problem is equivalent to optimizing the augmented Lagrangian $\mathcal{L}(\bm{\Theta}) = \sum_{i \in \mathcal{V}}\mathcal{L}_{i}(\bm{\Theta}_{i})$,
\small
\begin{align}
\mathcal{L}_{i}(\bm{\Theta}_{i}) &= 
	f_i(\textbf{$\theta_i$}) + \sum_{j \in \mathcal{B}_{i}} \left\{ \lambda^\top_{ij1} (\theta_i - \rho_{ij}) + \lambda^\top_{ij2}(\rho_{ij} - \theta_j) \right\} + \frac{\eta}{2} \sum_{j \in \mathcal{B}_{i}} \left\{ \| \theta_i - \rho_{ij} \|^2 + \| \rho_{ij} - \theta_j \|^2 \right\},
\label{eq:lagrangian}
\end{align}
\normalsize
where $\bm{\Theta} = \{\bm{\Theta}_{i}: i \in \mathcal{V}\}$, $\bm{\Theta}_{i} = \{\theta_{i}, \rho_{i}, \lambda_{i}\}$ are parameters to find, $\lambda_{i} = \{\lambda_{ij1}, \lambda_{ij2}: j \in \mathcal{B}_i\}$, $\lambda_{ij1}\text{, }\lambda_{ij2}$ are Lagrange multipliers, $\mathcal{B}_{i} = \{j | e_{ij} \in \mathcal{E} \} $ is the set of one hop neighbors of node $i$, $\eta > 0$ is a fixed scalar penalty constraint, and $\|\cdot\|$ is induced norm. The ADMM approach suggests that the optimization can be done in coordinate descent fashion taking gradient of each variable while fixing all the others.

\subsection{Convergence Speed of ADMM}

The currently known convergence rate of ADMM is $O(1/T)$ where $T$ is the number of iterations~\cite{he2012}. Even though $O(1/T)$ is the best known bound, it has been observed empirically that ADMM converges faster in many applications. Moreover, the computation time per each iteration may dominate the total algorithm running time. Thus many speed up techniques for ADMM have been proposed that are application specific. One way is to come up with a predictor-corrector step for the coordinate descent~\cite{goldstein2014} using some available acceleration method such as~\cite{nesterov1983}. It guarantees quadratic convergence for strongly convex $f_{i}(\cdot)$. 
Another way is to replace the gradient descent optimization with a stochastic one~\cite{ouyang2013, suzuki2013}. This approach has recently gained attention as it greatly reduces the computation per iteration. 
However, these methods usually require the coordinating center node thus may not readily applicable to the decentralized setting. Moreover, we want to preserve the application range of ADMM and avoid introducing additional assumptions on $f_{i}(\cdot)$.

One way to improve convergence speed of ADMM is through the use of different constraint penalty in each iteration. For example,~\cite{he2000} proposed ADMM with self-adaptive penalty, and it improved the convergence speed as well as made its performance less dependent on initial penalty values. The idea of \cite{he2000} is to change the constraint penalty taking account of the relative magnitudes of \emph{primal} and \emph{dual} residuals of ADMM as follows
\begin{align}
\eta^{t+1} = \left\{
	\begin{array}{ll}
	\eta^{t} \cdot (1+\tau^{t}) 		&\text{, if }{\| r^{t} \|}_2 > \mu {\| s^{t} \|}_2 \\[0.5em]
	\eta^{t} \cdot (1+\tau^{t})^{-1} 	&\text{, if }{\| s^{t} \|}_2 > \mu {\| r^{t} \|}_2 \\[0.5em]
	\eta^{t}                  			&\text{, otherwise }\\
	\end{array}
\right.
\label{eq:he2000}
\end{align}
where $t$ is the iteration index, $\mu > 1$, $\tau^{t} > 0$ are parameters, $r^{t}$ and $s^{t}$ are the primal and dual residuals, respectively\footnote{Please refer~\cite{boyd2010}, page 18 and 51 for their definitions.}. The primal residual measures the violation of the consensus constraints and the dual residual measures the progress of the optimization in the dual space. This update converges when $\tau^{t}$ satisfies $\sum_{t = 0}^{\infty} \tau^{t} < \infty$, i.e. we stop updating $\eta^{t}$ after a finite number of iterations. Typical choice for parameters are suggested as $\mu = 10$ and $\tau^{t} = 1$ at all $t$ iterations. The strength of this approach is that conservative changes in the penalty are guaranteed to converge~\cite{rockafellar1976, boyd2010}. 
However, like other ADMM speed up approaches mentioned above, this update scheme relies on the global computation of the primal and the dual residuals and requires the $\eta^{t}$ stored in nodes to be homogeneous over entire network thus it is not a fully decentralized scheme. Moreover, the choice of parameters as well as the maximum number of iterations require manually tuning.


\section{Proposed Methods}

We present our proposed ADMM penalty update schemes in three steps. First, we extend the aforementioned update scheme of (\ref{eq:he2000}) to be applicable on fully decentralized setting. Next, we propose the novel penalty parameter update strategy for ADMM speed up that does not require manual tuning of $\tau^{t}$. Finally, we extend the strategy so that we can automatically select the maximum number of penalty update iterations.

\subsection{ADMM with Varying Penalty (ADMM-VP)}

Throughout the paper, the superscript $t$ in all terms with subscript $i$ denote either the objective function or parameter at $t$-th iteration for node $i$. In order to extend (\ref{eq:he2000}) for a fully distributed setting, we first introduce $\eta_{i}^{t}$, the penalty for $i$-th node at $t$-th iteration. Next, we need to compute local primal and dual residuals for each node $i$. In the fully distributed learning framework of~\cite{forero2011, yoon2012}, the dual auxiliary variable vanishes from derivation. However, to compute the residuals, we need to keep track of the dual variable, which is essentially the average of local estimates, explicitly over iterations. The squared residual norms for the $i$-th node are defined as
\begin{align}
\| r_{i}^{t} \|_{2}^{2} = \| \theta_{i}^{t} - \bar{\theta}_{i}^{t} \|_{2}^{2}, \quad \| s_{i}^{t} \|_{2}^{2} = (\eta_{i}^{t})^{2} \| \bar{\theta}_{i}^{t} - \bar{\theta}_{i}^{t-1} \|_{2}^{2}, \quad \bar{\theta}_{i}^{t} = \frac{1}{|\mathcal{B}_{i}|} \sum_{j \in \mathcal{B}_{i}} \theta_{j}^{t}.
\end{align}
Note the difference from the standard residual definitions for consensus ADMM~\cite{boyd2010}, used in (\ref{eq:he2000}), where the dual variable is considered as a single, globally accessible variable, $\bar{\theta}^{t}$ instead of local $\bar{\theta}_{i}^{t}$. This allows each node to change its $\eta_{i}^{t}$ based on its own local residuals. The penalty update scheme is similar to (\ref{eq:he2000}) but $\eta^{t}$, $\| r^{t} \|_{2}$ and $\| s^{t} \|_{2}$ are replaced with $\eta_{i}^{t}$, $\| r_{i}^{t} \|_{2}$ and $\| s_{i}^{t} \|_{2}$, respectively. Lastly,~\cite{he2000} stopped changing $\eta^{t}$ after $t > 50$. However, in ADMM-VP, if we stop the same way, we end up with heterogeneously fixed penalty values which impacts the convergence of ADMM by yielding heavy oscillations near the saddle point. Therefore we reset all penalty values in all nodes to a pre-defined value (e.g. $\eta^{0}$, the initial penalty parameter) after a fixed number of iterations. As we fix the penalty values homogeneously after a finite number of iterations, it becomes the standard ADMM after that point thus the convergence of ADMM-VP update is guaranteed.

\subsection{ADMM with Adaptive Penalty (ADMM-AP)}
We further extend $\eta_{i}$ by introducing a bi-directional graph with a penalty constraint parameter $\eta_{ij}$ specific to directed edge $e_{ij}$ from node $i$ to $j$. The modified augmented Lagrangian $\mathcal{L}_{i}$ is similar to (\ref{eq:lagrangian}) except that we replace $\eta$ with $\eta_{ij}$.
The penalty constraint controls the amount each constraint contributes to the local minimization problem. The penalty constraint parameter $\eta_{ij}$ is determined by evaluating the parameter $\theta_{j}$ from node $j$ with the objective function $f_{i}(\cdot)$ of node $i$ as
\begin{flalign}
\eta^{t+1}_{ij} = \left\{
\begin{array}{ll} 
\eta^{0} \cdot (1+ \tau_{ij}^{t}) 		& \text{, if } t < t^{max} \\ [0.5em] 
\eta^{0}            					&\text{, otherwise}
\end{array}
\right. 
\label{eq:eta_update_ap}
\end{flalign}
where $t^{max}$ is the maximum number of iterations for the update as proposed in~\cite{he2000} and
\begin{align}
\tau_{ij}^{t}
	&= \frac{\kappa_{i}^{t}(\theta_{i}^{t})}{\kappa_{i}^{t}(\theta_{j}^{t})} - 1 \,, \quad
\kappa_{i}^{t}( \theta )
	=\left( \frac{ f_{i}^{t}(\theta) - f_{i}^{min} }{ f_{i}^{max} - f_{i}^{min} } + 1 \right)\,, \\
f_{i}^{max} 
	&= \max \{ f_{i}^{t}(\theta_{i}^{t}), f_{i}^{t}(\theta_{j}^{t}) : j \in \mathcal{B}_{i} \}\,, \quad
f_{i}^{min} 
	= \min \{ f_{i}^{t}(\theta_{i}^{t}), f_{i}^{t}(\theta_{j}^{t}) : j \in \mathcal{B}_{i} \}\,.
\end{align}
The interpretation of this update strategy is straightforward. In each iteration $t$, each $i$-th node will evaluate its objective using its own estimate of $\theta_{i}^{t}$ and the estimates from other nodes $\theta_{j}^{t}$ (we use $\rho_{ij}^{t}$ instead of actual $\theta_{j}^{t}$ to retain locality of each node from the neighbors). Then, we assign more weight to the neighbor with better parameter estimate for the local $f_{i}(\cdot)$ (i.e. larger penalty $\eta_{ij}^{t}$ if $f_{i}(\theta_{j}) < f_{i}(\theta_{i})$) with the above update scheme. The intuition behind the ADMM-AP update is to emphasize the local optimization during early stages and then deal with the consensus update at later, subsequence stages. If all local parameters yield similarly valued local objectives $f_{i}(\cdot)$, the onus is placed on consensus. This makes ADMM-AP different from pre-initialization that does the local optimization using the local observations and ignores the consensus constraints.


Note that unlike the update strategy of~(\ref{eq:he2000}), we do not need to specify $\tau^{t}$ and the update weight is automatically chosen according to the normalized difference in the local objective evaluation among neighboring parameters. The proposed algorithm also emphasizes the objective minimization over the minimization that solely depends on the norms of primal and dual residuals of constraints. The hope is that we not only achieve the consensus of the parameters of the model but also a \emph{good} estimate with respect to the objective.

On the other hand, the convergence property of~\cite{he2000} still holds for the proposed algorithm. Following Remark 4.2 of~\cite{he2000}, the requirement for the convergence is to satisfy the update ratio to be fixed after some $t^{\max} < \infty$ iteartions.
Moreover, the proposed update ensures bounding by $\eta_{ij}^{t+1} / \eta_{ij}^{t} \in [0.5, 2]$, which matches with the increase and decrease amount suggested in~\cite{he2000, boyd2010}. One may use $t^{\max} = 50$ as in~\cite{he2000}.

\subsection{ADMM with Network Adaptive Penalty (ADMM-NAP)}

To extend the proposed method for automatically deciding the maximum number of penalty updates, 
the penalty update for the ADMM becomes
\begin{flalign}
\eta^{t+1}_{ij} = \left\{
\begin{array}{ll} 
\eta^{0} \cdot (1+ \tau_{ij}^{t} ) 		& \text{, if } \sum_{u=1}^{t} |\tau_{ij}^{u}| < \mathcal{T}_{ij}^{t} \\ [0.5em] 
\eta^{0}            					&\text{, otherwise}.
\end{array}
\right. 
\label{eq:eta_update_nap}
\end{flalign}
Fig.~\ref{fig1:proposed} depicts how the proposed model have different structures from centralized and traditional distributed models, and how nodes share their parameters via network.

In addition to the adaptive penalty update, the inequality condition on the summation of $\tau_{ij}^{u}, u = 1..t$ encodes the spent \emph{budget} that the edge $e_{ij}$ can change $\eta_{ij}$. All nodes have its upper bound $\mathcal{T}_{ij}^{t}$ and everytime it makes a change to $\eta_{ij}$, it has to \emph{pay} exactly the amount they changed. If the edge has changed too much, too often, the update strategy will block the edge from changing $\eta_{ij}$ any more.



The update scheme is guaranteed to convergence if $\mathcal{T}_{ij}^{t}$ is simply set to constant $\mathcal{T}$ for all $i, j, t$ or if $\tau_{ij}^{t} = 0$ for $t > t^{max}$. However, with a different objective function and different network connectivity, a different upper bound should be imposed. This is because a given upper bound $\mathcal{T}$ or maximum iteration $t^{max}$ could be too small for a certain node to fully take an advantage of our adaptation strategy or they could be too big so that it converges much slowly because of the continuously changing $\eta_{ij}^{t}$. To this end, we propose updating strategy for $\mathcal{T}_{ij}^{t}$ as following:
\begin{flalign}
\mathcal{T}_{ij}^{t+1} = &\left\{
\begin{array}{ll}
	\mathcal{T}_{ij}^{t} + \alpha^{n} \mathcal{T} 			&\text{, if }
	\sum_{u=1}^{t} |\tau_{ij}^{u}| \geq \mathcal{T}_{ij}^{t} 
	~~\text{and } | f_{i}(\theta_{i}^{t}) - f_{i}(\theta_{i}^{t-1}) | > \beta \\ [1em]
	\mathcal{T}_{ij}^{t} 								& \text{, otherwise }\\
\end{array}
\right.
\label{eq:T_update}
\end{flalign}
where $\mathcal{T}_{ij}^0$ is set by an initial parameter $\mathcal{T}$ and $\alpha, \beta \in (0, 1)$ are parameters. Whenever $\mathcal{T}_{ij}^{t+1} > \mathcal{T}_{ij}^{t}$, we increase $n$ by 1. Once $\sum_{u=1}^{t} |\tau_{ij}^{u}| \geq \mathcal{T}_{ij}^{t}$ but its objective value is still significantly changing, i.e. $| f_{i}(\theta_{i}^{t}) - f_{i}(\theta_{i}^{t-1}) | > \beta$, 
$\mathcal{T}_{ij}^{t+1}$ is increased by $\alpha^n \mathcal{T}$. Note that the independent upper bound $\mathcal{T}_{ij}^{t}$ for each $\eta_{ij}^{t}$ update on the edge $e_{ij}$ makes it sensitive to the various network topology, but it still satisfies the convergence condition because
\begin{flalign}
\lim_{t \rightarrow \infty} \mathcal{T}_{ij}^{t} 
	\leq \sum_{n=1}^{\infty} \alpha^{n-1}~\mathcal{T}
	= \frac{1}{1-\alpha}\mathcal{T}.
\end{flalign}


\subsection{Combined Update Strategies (ADMM-VP + AP, ADMM-VP + NAP)}
Observing (\ref{eq:he2000}) and the proposed update schemes (\ref{eq:eta_update_ap}) and (\ref{eq:eta_update_nap}), one can easily come up with a combined update strategy by replacing $\tau^{t}$ in (\ref{eq:he2000}) with $\tau_{ij}^{t}$. Based on preliminary experiments, we found that this replacement yields little utility. Instead, we suggest another penalty update strategy combining ADMM-VP and ADMM-AP as
\begin{align}
\eta_{ij}^{t+1} = \left\{
	\begin{array}{ll}
	\eta_{ij}^{t} \cdot (1 + \tau_{ij}^{t}) \cdot 2		&\text{, if }{\| r_{i}^{t} \|}_2 > \mu {\| s_{i}^{t} \|}_2 \\[0.5em] 
	\eta_{ij}^{t} \cdot (1 + \tau_{ij}^{t}) \cdot (1/2)		&\text{, if }{\| s_{i}^{t} \|}_2 > \mu {\| r_{i}^{t} \|}_2 \\[0.5em] 
	\eta_{ij}^{t}                  							&\text{, otherwise }\\
	\end{array}
\right.
\label{eq:combined}
\end{align}
which we denote as ADMM-VP + AP. We reset $\eta_{ij}^{t} = \eta^{0}$ when $t > t^{\max}$. In order to combine ADMM-VP and ADMM-NAP, we consider the summation condition of $\tau_{ij}^{t}$ as in (\ref{eq:eta_update_nap}). We denote this strategy as ADMM-VP + NAP.

\section{Distributed Maximum Likelihood Learning}
\label{sec:dpl}

In this section, we show how our method can be applied to an existing distributed learning framework in the context of distributed probabilistic principal component analysis (D-PPCA). 
D-PPCA can be viewed as fundamental approach to a general matrix factorization task in the presence of potentially missing data, with many applications in machine learning.

\subsection{Probabilistic Principal Component Analysis}

The Probabilistic PCA (PPCA) \cite{tipping1999} has many applications in vision problems, including structure from motion, dictionary learning, image inpainting, etc. We here restrict our attention to the linear PPCA without any loss of generalization. The centralized PPCA is formulated as the task of projecting the source data $\mathbf{x}$ according to
$
\mathbf{x} = \mathbf{W} \mathbf{z} + \bm{\mu} + \bm{\epsilon}
$
where $\mathbf{x} \in \mathbb{R}^{D}$ is the observation column vector, $\mathbf{z} \in \mathbb{R}^{M}$ is the latent variable following $\mathbf{z} \sim\mathcal{N}(\mathbf{0},\mathbf{I})$, $\mathbf{W} \in \mathbb{R}^{D \times M}$ is the projection matrix that maps $\mathbf{x}$ to $\mathbf{z}$, $\bm{\mu} \in \mathbb{R}^{D}$ allows non-zero mean, and the Gaussian observation noise $\bm{\epsilon}\sim\mathcal{N}(\mathbf{0},a^{-1}\mathbf{I})$ with the noise precision $a$. When $a^{-1} = 0$, PPCA recovers the standard PCA. The posterior estimate of the latent variable $\mathbf{z}$ given the observation $\mathbf{x}$ is
\begin{flalign}
p( \mathbf{z}|\mathbf{x} ) \sim \mathcal{N}(\mathbf{M}^{-1}\mathbf{W}^\top ( \mathbf{x}- \bm{\mu} ), a^{-1} \mathbf{M}^{-1}),
\label{eq:posterior_zx}
\end{flalign}
where $\mathbf{M}=\mathbf{W}^\top\mathbf{W}+a^{-1}\mathbf{I}$. The parameters $\mathbf{W}$, $\bm{\mu}$, and $a$ can be estimated using a number of methods, including SVD and Expectation Maximization (EM) algorithm.

\subsection{Distributed PPCA}

The distributed extension of PPCA (D-PPCA)~\cite{yoon2012} can be derived by applying ADMM to the centralized PPCA model above. Each node learns its local copy of PPCA parameters with its set of local observations $\mathbf{X}_i = \{ \mathbf{x}_{in} | n = 1..N_{i} \}$ where $\mathbf{x}_{in}$ denotes the $n$-th observation in $i$-th node and $N_{i}$ is the number of observations available in the node. Then, they exchange the parameters using the Lagrange multipliers and impose consensus constraints on the parameters. The global constrained  optimization is
\begin{align}
\label{eq:obj}
\min_{\bm{\Theta}_i}\, -\log p(\mathbf{X}_i | \bm{\Theta}_i)\quad s.t. &\quad \bm{\Theta}_i = \rho_{ij}^{\bm{\Theta}}, \rho_{ij}^{\bm{\Theta}} = \bm{\Theta}_j, 
\end{align}
where $i \in \mathcal{V}, j \in \mathcal{B}_i$, $\bm{\Theta}_i = \{\mathbf{W}_i, \bm{\mu}_i, a_i\}$ is the set of local parameters and $\rho_{ij}^{\bm{\Theta}} = \{ \rho_{ij}^{\mathbf{W}}, \rho_{ij}^{\bm{\mu}}, \rho_{ij}^{a} \}$ is the set of auxiliary variables for the parameters. For the details regarding how the decentralized model is optimized, see~\cite{yoon2012}.

\vspace{-0.1in}
\subsection{D-PPCA with Network Adpative Penalty}

The augmented Lagrangian applying the proposed ADMM with Network Adpative Penalty is similar to~\cite{yoon2012} except that $\eta$ becomes $\eta_{ij}$.
with $\lambda_{i}$, $\gamma_{i}$, $\beta_{i}$ are Lagrange multipliers for the PPCA parameters for node $i$. 
The adaptive penalty constraint $\eta_{ij}^{t}$ controls the speed of parameter propagation dynamically so that the overall optimization empirically converges faster than~\cite{yoon2012}. 
One can solve this optimization using the distributed EM approach~\cite{forero2011}. The E-step of the D-PPCA is the same as centralized counterpart~\cite{tipping1999}. The M-step is similar to~\cite{yoon2012} except we use separate $\eta_{ij}$ for each edge. Since the update formulas for the three parameters are similar, we present the $\bm{\mu}_{i}$ update as an example. First, $\bm{\mu}_{i}$ can be updated as
{\small
\begin{flalign}
\label{eq:mu}
&\bm{\mu}_i^{t+1} = \left\{ 
	a_i \sum_{n=1}^{N_i}\left( \textbf{x}_{in} - \textbf{W}_i \mathbb{E}[\textbf{z}_{in}] \right) - 2\gamma_i^{t} + \sum_{j \in \mathcal{B}_i} \eta_{ij} \left( \bm{\mu}_i^{t} + \bm{\mu}_j^{t} \right) \right\} \cdot \left( N_i a_i + 2\sum_{j \in \mathcal{B}_i} \eta_{ij}^{t} \right)^{-1},
\end{flalign}
}%
where $\mathbb{E}[\mathbf{z}_{in}]$ denotes the posterior estimates of the $n$-th latent variable of node $i$. Note that unlike D-PPCA where we computed the normalization factor as $N_{i} a_{i} + 2 \eta | \mathcal{B}_{i} |$ where $| \cdot |$ is the cardinality, we add up $\eta_{ij}^{t}, \forall j \in \mathcal{B}_{i}$. The corresponding Lagrange multiplier can be computed as penalty-weighted summation of consensus errors
$\gamma_i^{t+1} = \gamma_i^{t} + (1/2)\sum_{j \in \mathcal{B}_i} \eta_{ij}^{t} \left( \bm{\mu}_i^{t+1} - \bm{\mu}_j^{t+1} \right)$.
Once all the parameters and the Lagrange multipliers are updated, we update $\eta_{ij}$ and $\mathcal{T}_{ij}$ using (\ref{eq:eta_update_nap}) and (\ref{eq:T_update}), respectively. Algorithm 1 in the appendix summarizes the overall steps for the D-PPCA with Network Adpative Penalty.

\section{Experiments}

We first analyze and compare the proposed methods (ADMM-VP, ADMM-AP, ADMM-NAP, ADMM-VP + AP, ADMM-VP + NAP) with the baseline method using synthetic data. Next, we apply our method to a distributed structure from motion problem using two benchmark real world datasets. For the baseline, we compare with the standard ADMM-based D-PPCA~\cite{yoon2012} denoted as \texttt{ADMM}. 
Unless noted otherwise, we used $\eta^{0} = 10$. To assess convergence, we compare the relative change of~(\ref{eq:obj}) to a fixed threshold ($10^{-3}$ in this case) for the D-PPCA experiments as in~\cite{yoon2012}.

\subsection{Synthetic Data}

\begin{figure*}[t]
\centering
$\begin{array}{c c c}
	\begin{subfigure}{0.31\textwidth}
		\includegraphics[width=1\textwidth]{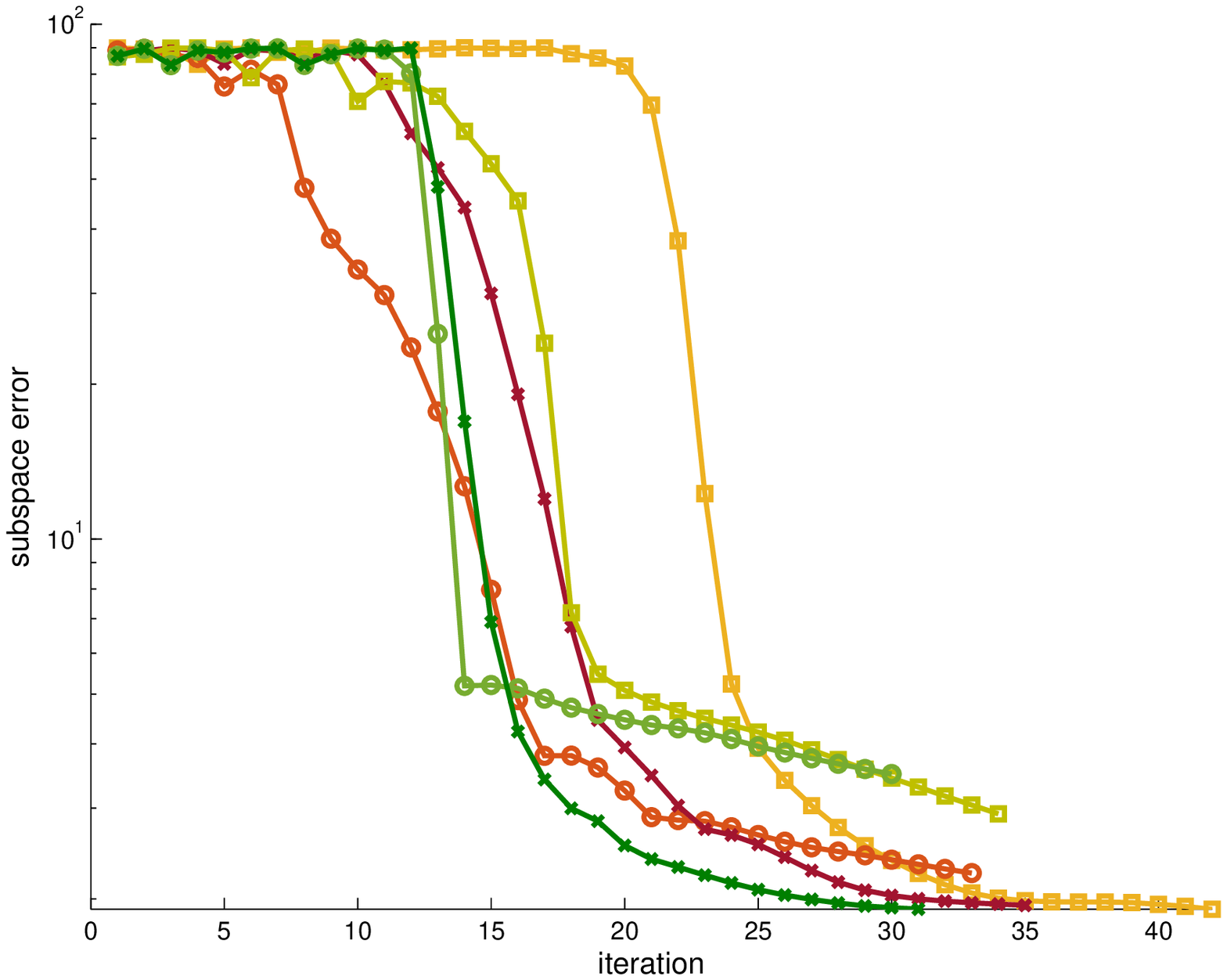}
		\caption{12 nodes (complete)}
		\label{fig_s1}
	\end{subfigure}
	&
	\begin{subfigure}{0.31\textwidth}
		\includegraphics[width=1\textwidth]{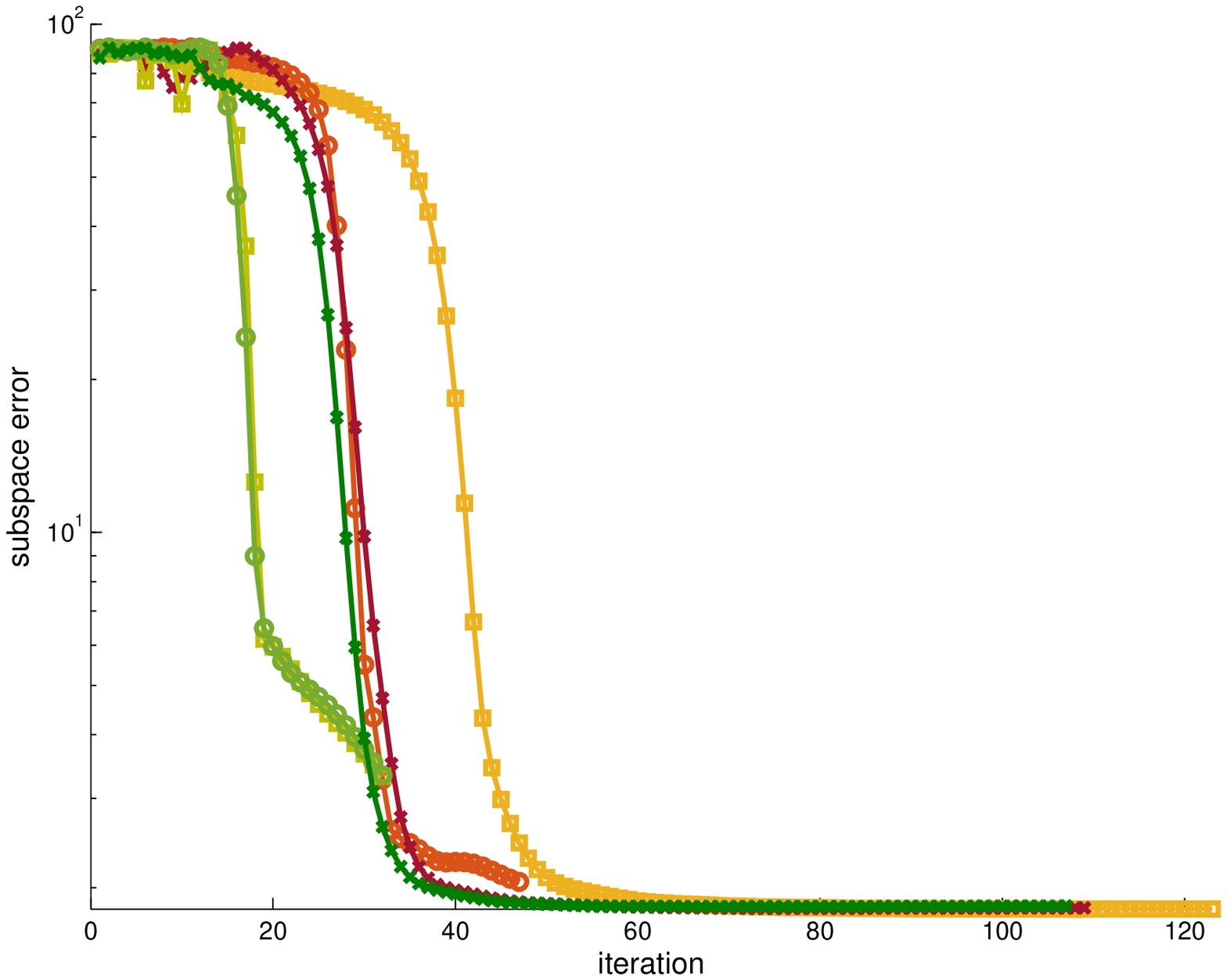}
		\caption{16 nodes (complete)}
		\label{fig_s2}
	\end{subfigure}
	&
	\begin{subfigure}{0.31\textwidth}
		\includegraphics[width=1\textwidth]{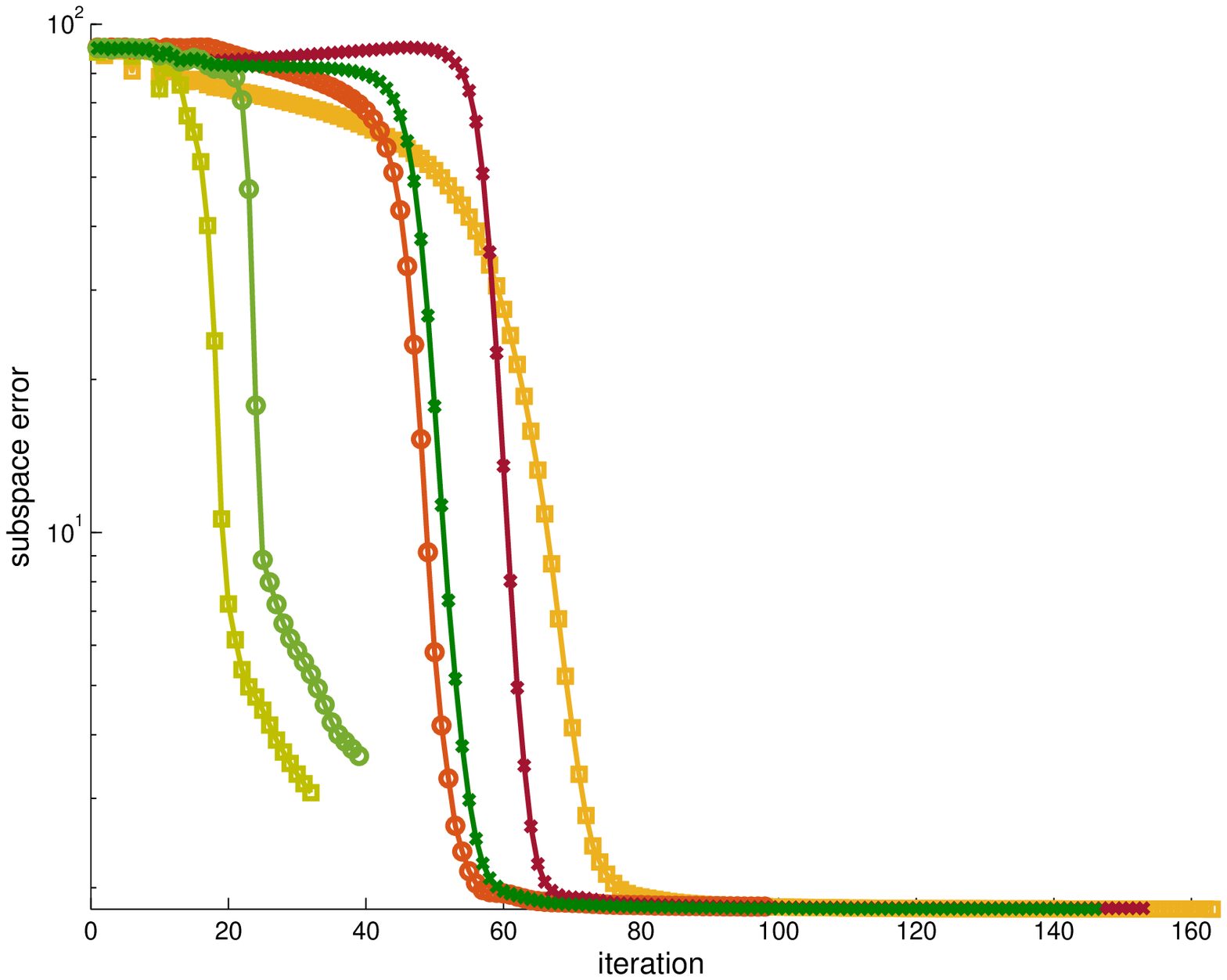}
		\caption{20 nodes (complete)}
		\label{fig_s3}
	\end{subfigure}
	\\
	\begin{subfigure}{0.31\textwidth}
		\includegraphics[width=1\textwidth]{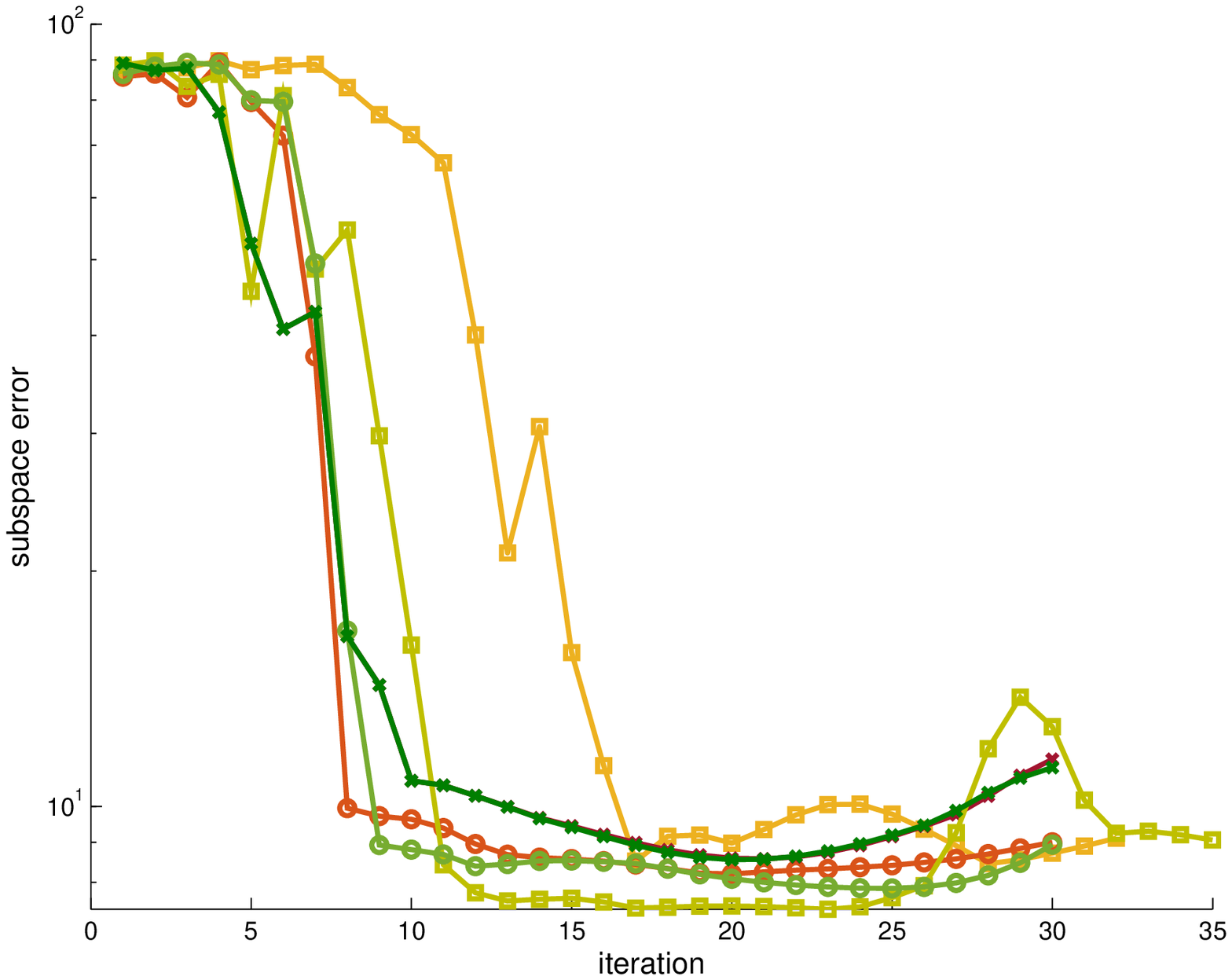}
		\caption{20 nodes (ring)}
		\label{fig_s4}
	\end{subfigure}
	&
	\begin{subfigure}{0.31\textwidth}
		\includegraphics[width=1\textwidth]{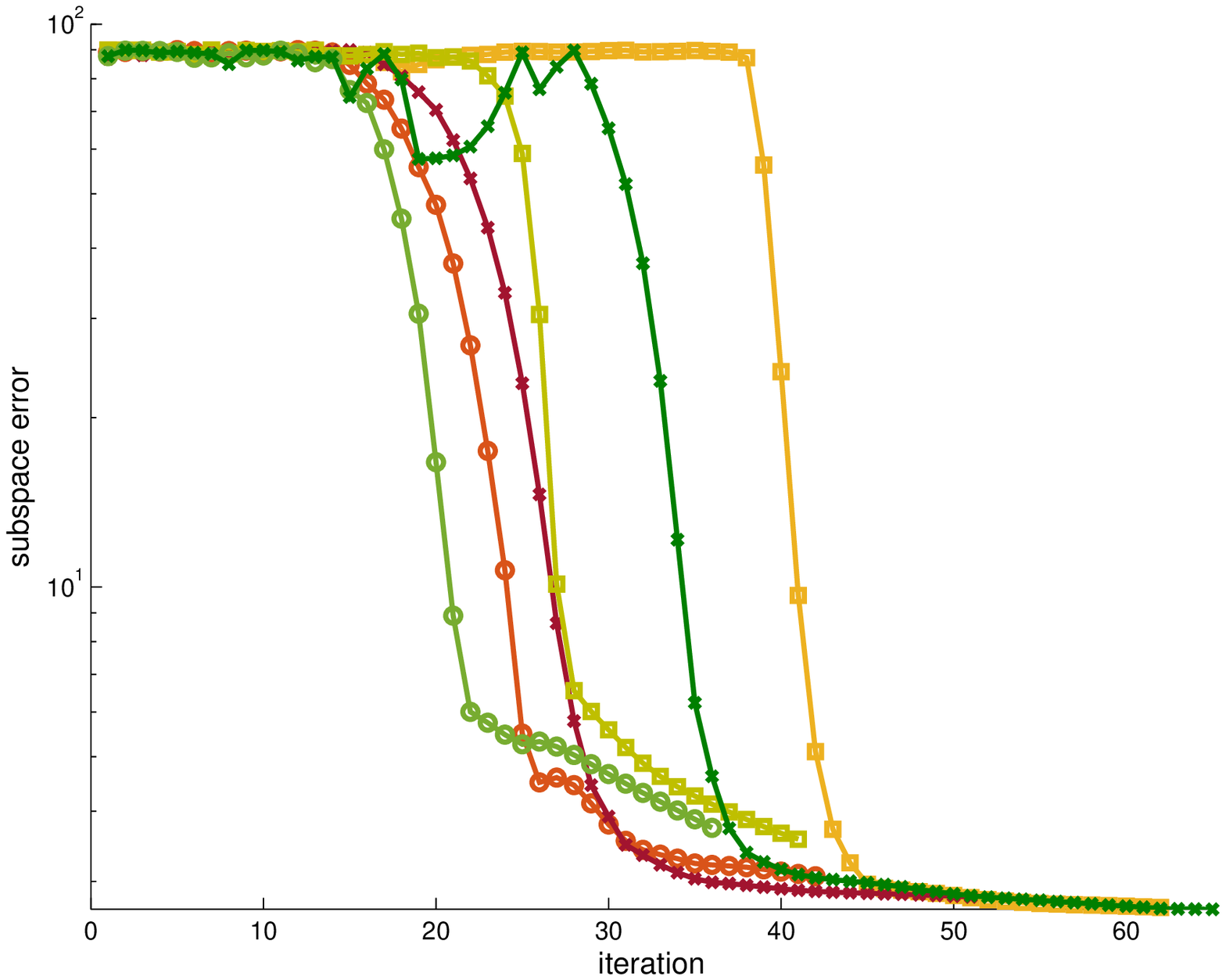}
		\caption{20 nodes (cluster)}
		\label{fig_s5}
	\end{subfigure}
	&
	\begin{subfigure}{0.25\textwidth}
		\includegraphics[width=1\textwidth]{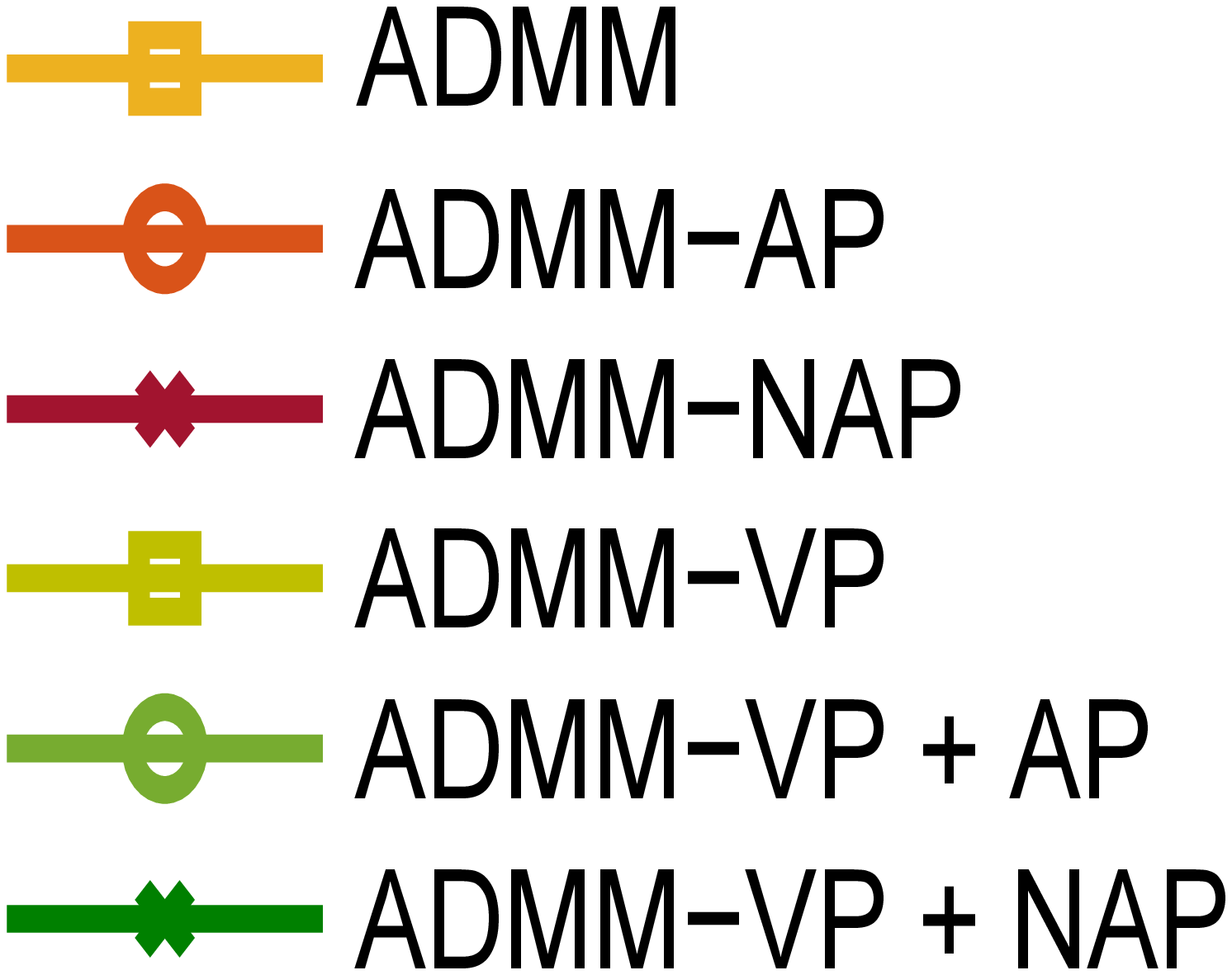}
		\label{fig_s6}
	\end{subfigure}
	\\
	\\
\end{array}$
	\caption{The comparison of proposed methods and the baseline ADMM using the subspace angle error of the projection matrix with (a-c) different graph size and (c-e) different network topology}
	\label{fig:synthetic1}
\vspace{-0.2in}
\end{figure*}

We generated 500 samples of 20 dimensional observations from a 5-dim subspace following $\mathcal{N}(\mathbf{0}, \mathbf{I})$, with the Gaussian measurement noise following $\mathcal{N}(\mathbf{0}, 0.2 \cdot \mathbf{I})$. For the distributed settings, the samples are assigned to each node evenly. All experiments are ran with 20 independent random initializations. We measured the number of iterations to convergence and the maximum subspace angle error versus the ground truth defined as the maximum of subspace angles between each node's projection matrix and the ground truth projection matrix. We examined the impact of different graph topologies and different graph sizes. We tested three network topologies: complete, ring and cluster (a connected graph consists of two complete graphs linked with an edge). For the graph size, we tested on 12, 16 and 20 nodes settings.

Top three plots in Fig.~\ref{fig:synthetic1} depict results over varying number of nodes while fixing the graph topology as the complete graph. We plot the median result out of the 20 independent initializations. We observed that the speed up with the proposed method, particularly for ADMM-VP and its variants, becomes more significant as the number of nodes increases. This suggests the proposed method can be of particular use as the size of an application problem increases. Fig.~\ref{fig_s3} to Fig.~\ref{fig_s5} in the figure show the performance in the context of different network topologies. Our proposed methods converge faster or at the same rate as the standard ADMM. The proposed method works most robustly in the complete graph setting. In other words as the graph connectivity increases, the convergence property of the proposed method improves. Note also that ADMM-VP works best in complete graph while ADMM-AP / NAP are better than the ADMM-VP in weakly connected networks. This makes sense as ADMM-VP depends on residual computation and the proposed local residual computation become less accurate compared to the complete graph when the global residual can be computed.

%
%
%
%

\subsection{Distributed Affine Structure from Motion}

We tested the performance of our method on five objects of Caltech Turntable~\cite{Moreels2007} and Hopkins 155~\cite{tron2007} dataset as in~\cite{yoon2012}. The goal here is to jointly estimate the 3D structure of the objects as well as the camera motion, however in a distributed camera network setting. The input measurement matrix is defined as $2 \times F$ by $N$ where $F$ denotes the number of frames and $N$ denotes the number of points. By applying PCA, we can decompose the input into the camera pose $\mathbf{W}_{i}$ and the 3D structure $\mathbb{E}[\mathbf{z}_{in}], n = 1..N_{i}$. For the detailed experimental setting, refer to~\cite{tron2011, yoon2012}. As the performance measure, we used the maximum subspace angle error versus the centralized SVD-reconstructed structure. The network setting assumes five cameras on a complete graph.

\begin{figure*}[t]
\centering
$\begin{array}{c c c c c}
	\begin{subfigure}[h]{0.27\textwidth}
		\includegraphics[width=1\textwidth]{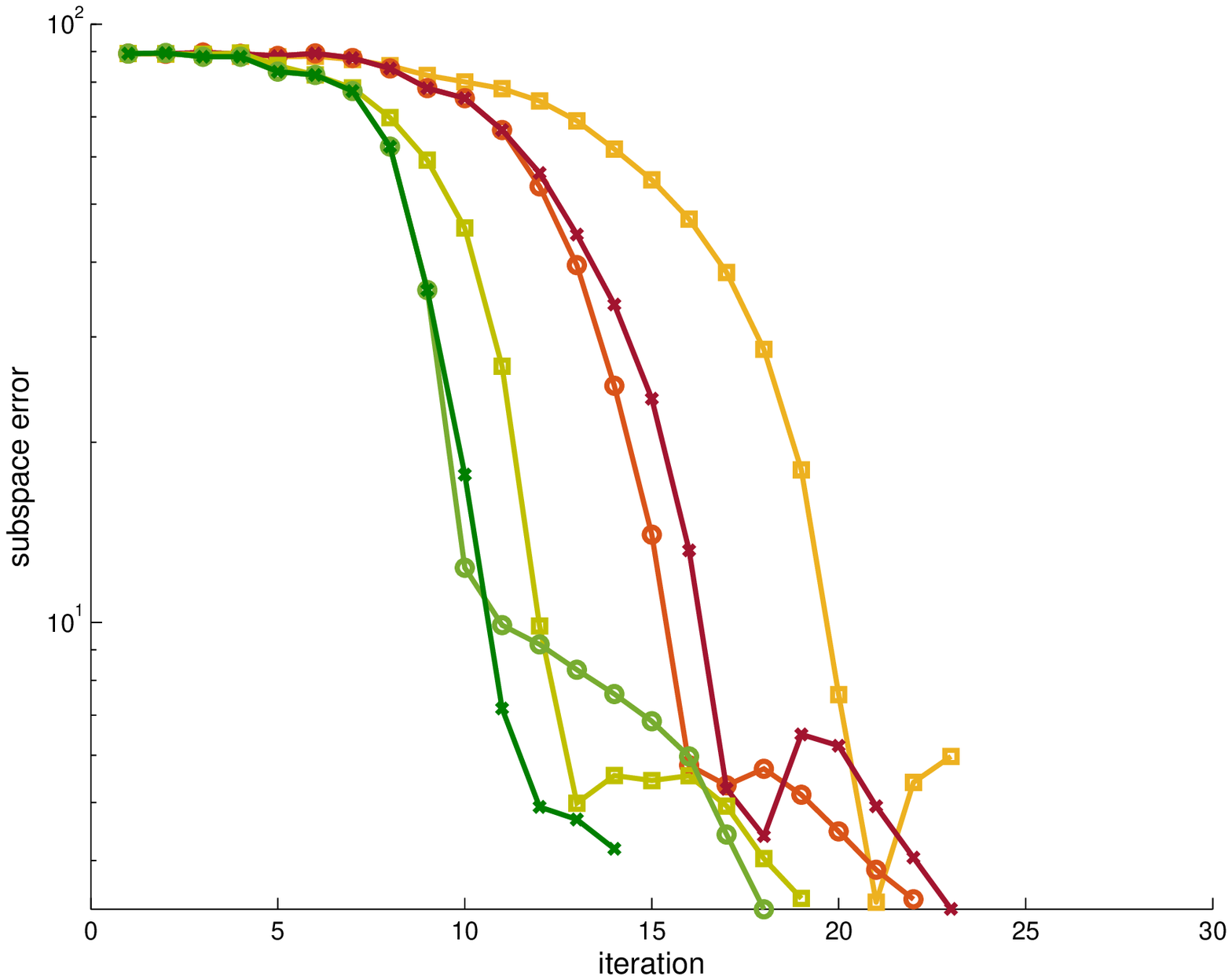}
		\caption{$t^{max} = 50$ (ring)}
		\label{fig_c1}
	\end{subfigure}
	&
	\begin{subfigure}[h]{0.27\textwidth}
		\includegraphics[width=1\textwidth]{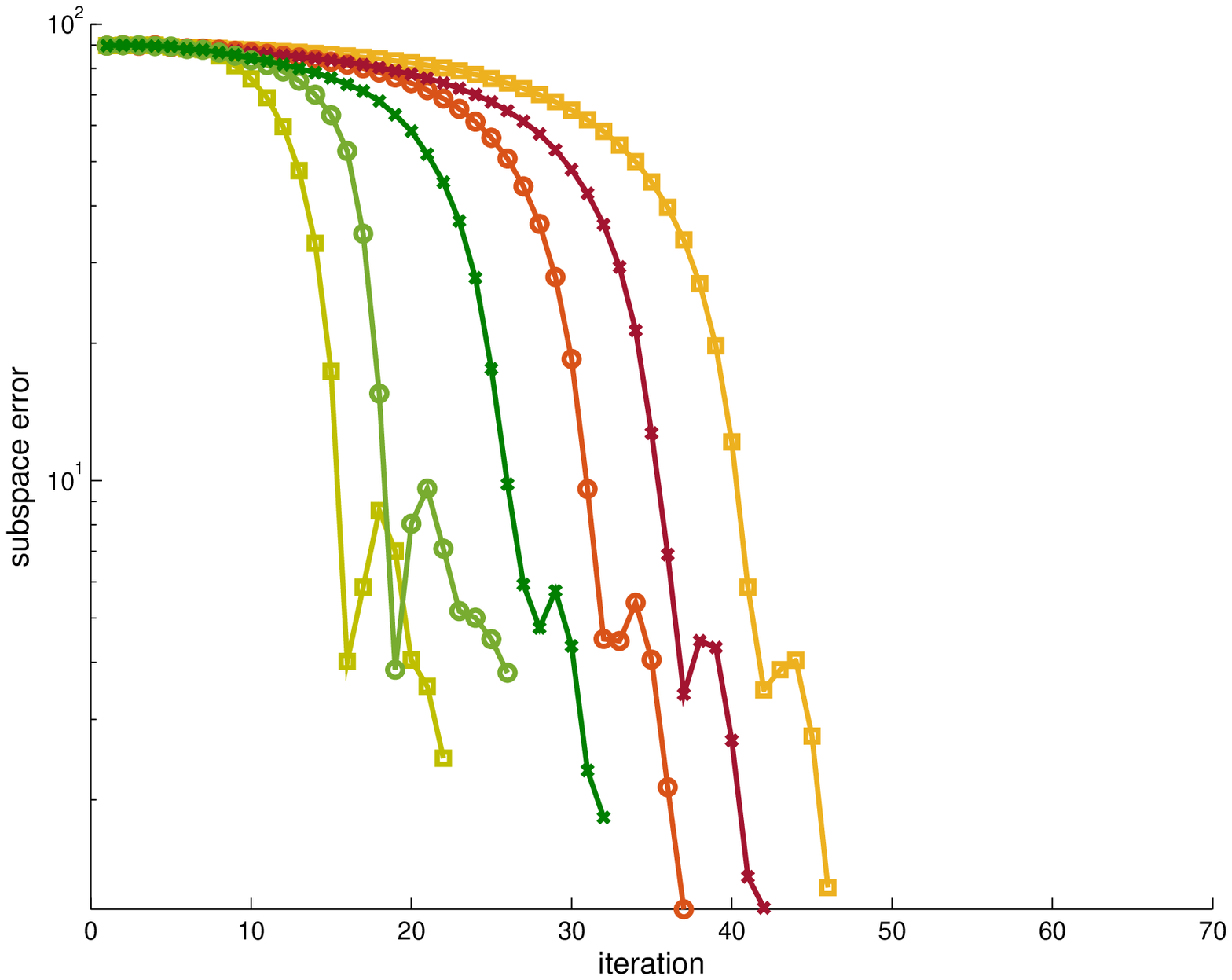}
		\caption{$t^{max} = 50$ (complete)}
		\label{fig_c2}
	\end{subfigure}
	&
	\begin{subfigure}[h]{0.27\textwidth}
		\includegraphics[width=1\textwidth]{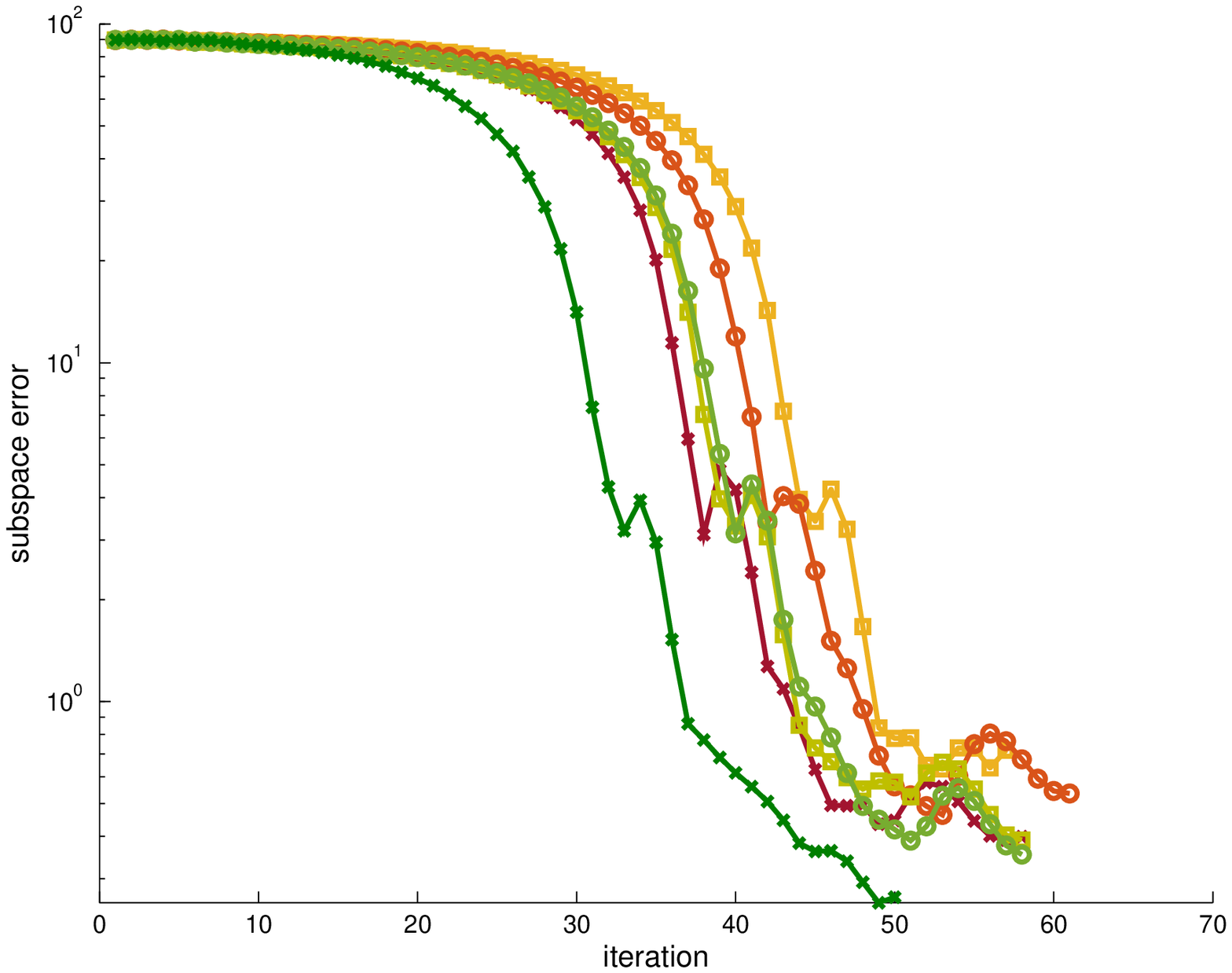}
		\caption{$t^{max} = 5$ (complete)}
		\label{fig_c3}
	\end{subfigure}
\end{array}$
	\caption{The comparison of proposed methods and the baseline ADMM using the subspace angle error of the reconstructed 3D structure with one object in Caltech dataset (Standing). Results on the remaining four objects can be found in the appendix. See Fig.~\ref{fig:synthetic1} for the plot labels.}
	\label{fig:caltech}
\vspace{-0.25in}
\end{figure*}

Fig.~\ref{fig:caltech} shows the result on the Caltech Turntable dataset. First, we compare Fig.~\ref{fig_c1} and Fig.~\ref{fig_c2}. One can see that when the graph is less connected (Fig.~\ref{fig_c1}), the proposed adaptive penalty method can boost ADMM-VP which cannot utilize the full residual information of fully connected case (Fig.~\ref{fig_c2}), as explained in synthetic data experiments. Next, we compare Fig.~\ref{fig_c2} and Fig.~\ref{fig_c3}. The network topologies are the same (complete) but $t^{\max}$ value required for ADMM-VP, ADMM-AP, ADMM-VP + AP is different in these two groups of experiments. When $t^{max}$ = 50 (Fig.~\ref{fig_c2}), all methods can accelerate throughout the iterations. However, when $t^{max}$ = 5 (Fig.~\ref{fig_c3}), the methods that depend on $t^{max}$ cannot accelerate after 5 iterations thus showing behavior similar to the baseline ADMM. On the other hand, ADMM-NAP based methods can accelerate by adaptively modifying the maximum number of penalty updates. Note that one can choose any small value of $\mathcal{T}$ and $\mathcal{T}_{ij}$ is increased automatically using (\ref{eq:T_update}).

For the Hopkins 155 dataset, we compared methods on 135 objects using the same approach as~\cite{yoon2012}. For each method considered, we computed the mean number of iterations until convergence. Since some objects in the dataset are point trajectories of non-rigid structure, it is inevitable for simple linear models to fail for those objects. Thus we omitted objects yielded more than 15 degrees when calculating the mean. For each object, we tested 5 independent random initializations. For ADMM-AP, ADMM-NAP and ADMM-VP + NAP, we found no significant speed up over the baseline ADMM. For ADMM-VP and ADMM-VP + AP, we could obtain 40.2\%, 37.3\% speed up, respectively if we use complete network. In ring network, the amount of improvement becomes smaller. This small or no improvement of speed is mainly due to the fact that the baseline ADMM converges fast enough (typically $< 100$ iterations) thus there is little room for the proposed methods to speed up the optimization. As observed from the synthetic experiments and Caltech dataset, the acceleration of the proposed methods occurs at the earlier iterations of the optimization. Thus if one can come up with a better convergence checking criterion depending on the application, the proposed methods can be a very viable choice due to its parameter-free nature.

\vspace{-0.1in}

\section{Conclusion}
We introduced a novel adaptive penalty update methods for ADMM that can be applied to consensus distributed learning frameworks. Contrary to previous approaches, our adaptive penalty update methods, ADMM-AP and ADMM-NAP does not depend on the parameters that require manual tuning. Using both synthetic and real data experiments, we showed the empirical effectiveness of the methods over the baseline. In addition, we found that the performance of ADMM-VP decreases with weakly connected graphs, and in those cases, ADMM-AP and ADMM-NAP can be useful.

The proposed methods do leave some room for improvements. For the problems when the standard ADMM can converge fast enough, the proposed methods may show less than significant gains. A better convergence criterion may help stop the proposed algorithms at earlier iterations (e.g. a criterion that can stop algorithms to remove long tails in Fig.~\ref{fig_s2} or Fig.~\ref{fig_s3}). 

\appendix

\section{D-PPCA with Network Adaptive Penalty}
Here we summarize the distributed probabilistic principal component analysis (D-PPCA)~\cite{yoon2012} algorithm modified to use the proposed network adaptive penalty update scheme (ADMM-NAP). We follow the notations from the previous sections.
The D-PPCA with Network Adaptive Penalty algorithm is summarized in Algorithm~\ref{algo:SDPPCA}.

\begin{algorithm}[!htbp]
\caption{D-PPCA with Network Adpative Penalty}\label{algo:SDPPCA}
\begin{algorithmic}[1]
\REQUIRE For every node $i$ randomly initialize $\textbf{W}_i^{0}, \bm{\mu}_i^{0}, a_i^{0}$
and set $\lambda_i^{0}=0, \gamma_i^{0}=0,\beta_i^{0}=0, \eta_{ij}=\eta$ for $j \in \mathcal{B}_i$
\FOR{$t=0,1,2,\cdots$ until convergence}
	\FORALL{$i \in \mathcal{V}$}
		\STATE Compute $\mathbb{E}[\textbf{z}_{in}]$ and $\mathbb{E}[\textbf{z}_{in}\textbf{z}_{in}^\top]$
		\STATE Compute $\textbf{W}_i^{t+1}, \bm{\mu}_i^{(t+1)}, a_i^{(t+1)}$
	\ENDFOR
	\FORALL{$i \in \mathcal{V}$}
		\STATE Broadcast $\textbf{W}_i^{(t+1)}, \bm{\mu}_i^{(t+1)},$ and $a_i^{(t+1)}$ to $\forall j \in \mathcal{B}_i$
	\ENDFOR
	\FORALL{$i \in \mathcal{V}$}
		\STATE Compute \textbf{$\lambda_i^{(t+1)}$}, \textbf{$\gamma_i^{(t+1)}$}, and $\beta_i^{(t+1)}$ 
	\ENDFOR	
	\FORALL{$i \in \mathcal{V}$}
		\STATE Update $\eta_{ij}$ for $j \in \mathcal{B}_i$ via (\ref{eq:eta_update_nap})
		\STATE Update $\mathcal{T}_{ij}$ for $j \in \mathcal{B}_i$ via (\ref{eq:T_update})
	\ENDFOR
\ENDFOR
\end{algorithmic}
\end{algorithm}

\section{Results on Caltech Turntable Dataset}

\begin{figure*}[b]
	\centering
	\begin{subfigure}[h]{0.49\textwidth}
		\includegraphics[width=1\textwidth]{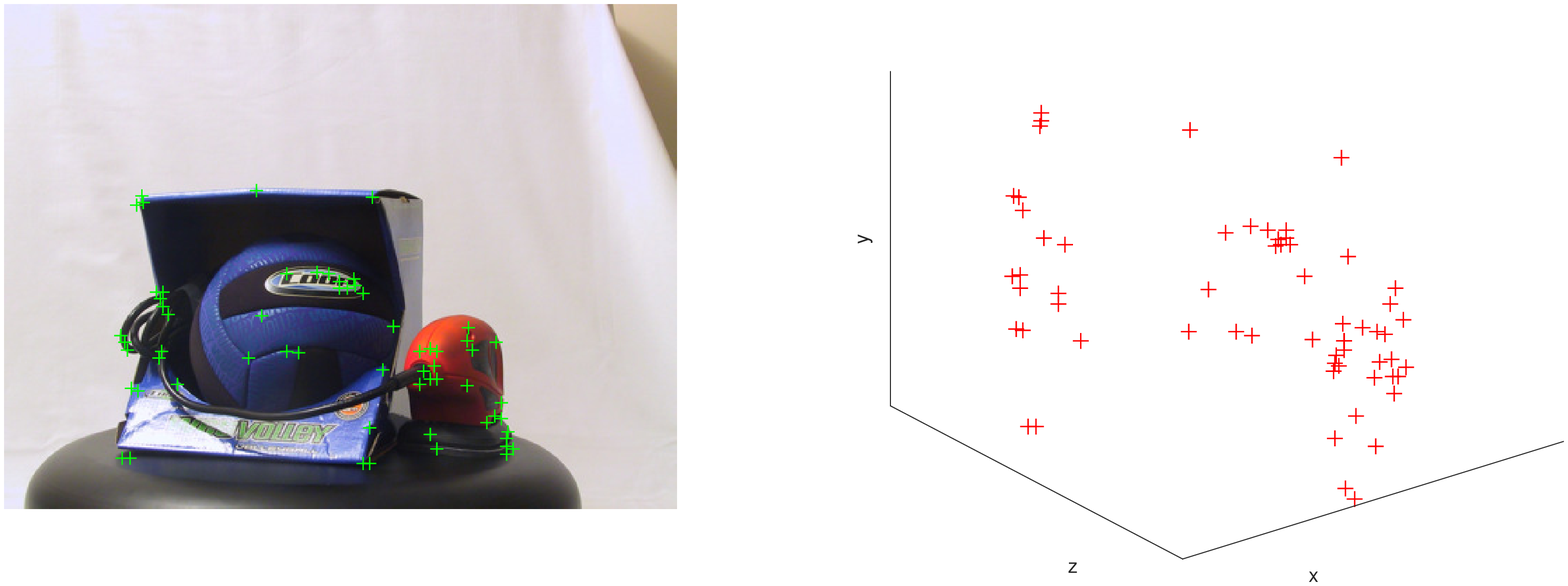}
		\caption{BallSander (62 points)}
		\label{fig0a}
	\end{subfigure}
	\begin{subfigure}[h]{0.49\textwidth}
		\includegraphics[width=1\textwidth]{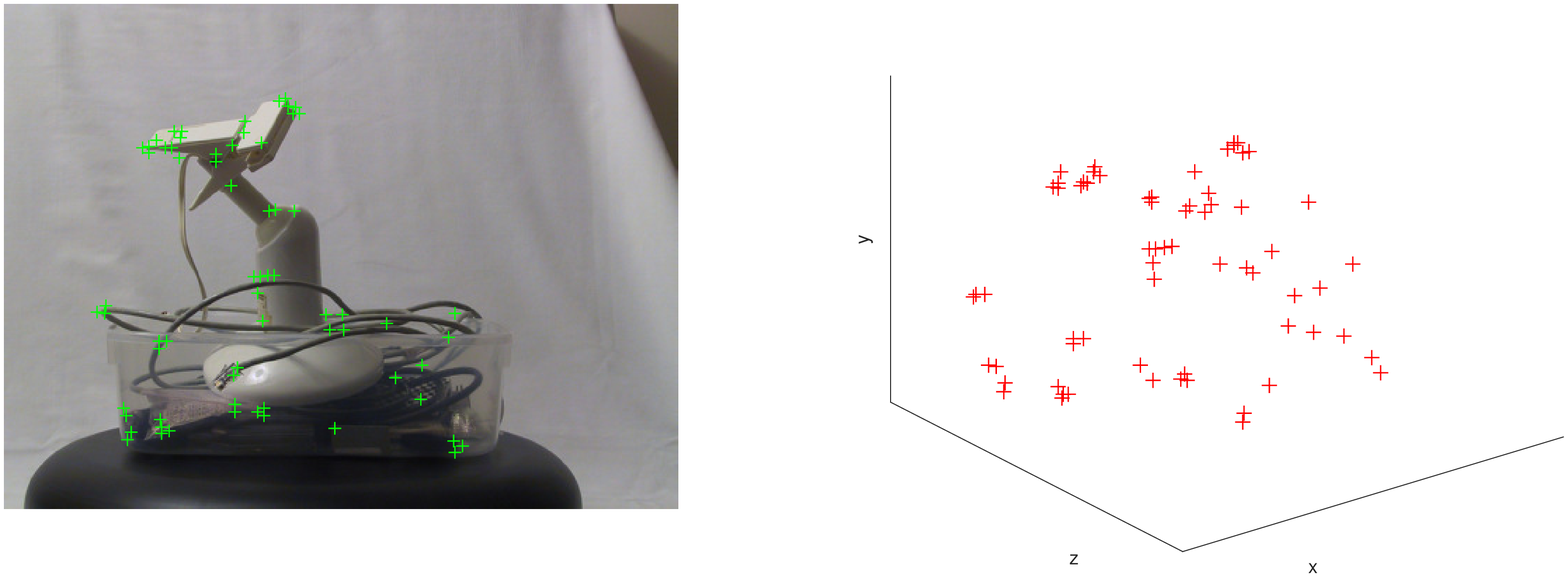}
		\caption{BoxStuff (67 points)}
		\label{fig0b}
	\end{subfigure}
	\\
	\begin{subfigure}[h]{0.49\textwidth}
		\includegraphics[width=1\textwidth]{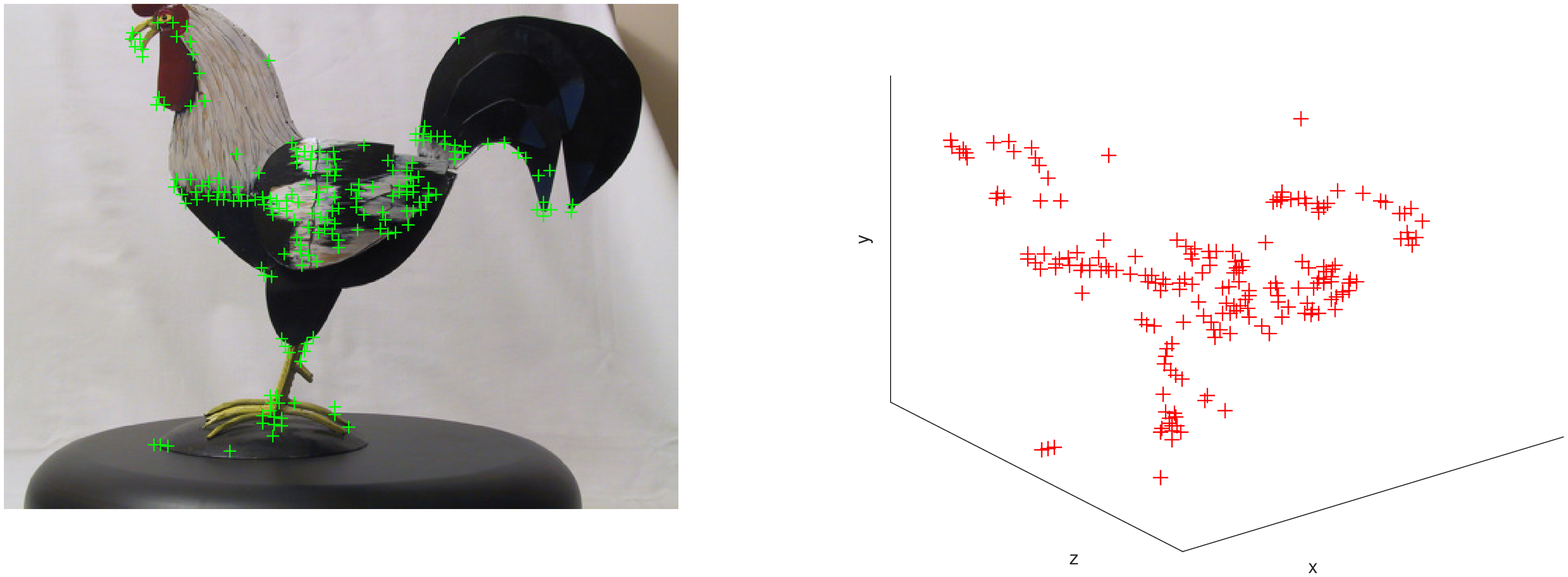}
		\caption{Rooster (189 points)}
		\label{fig0c}
	\end{subfigure}
	\begin{subfigure}[h]{0.49\textwidth}
		\includegraphics[width=1\textwidth]{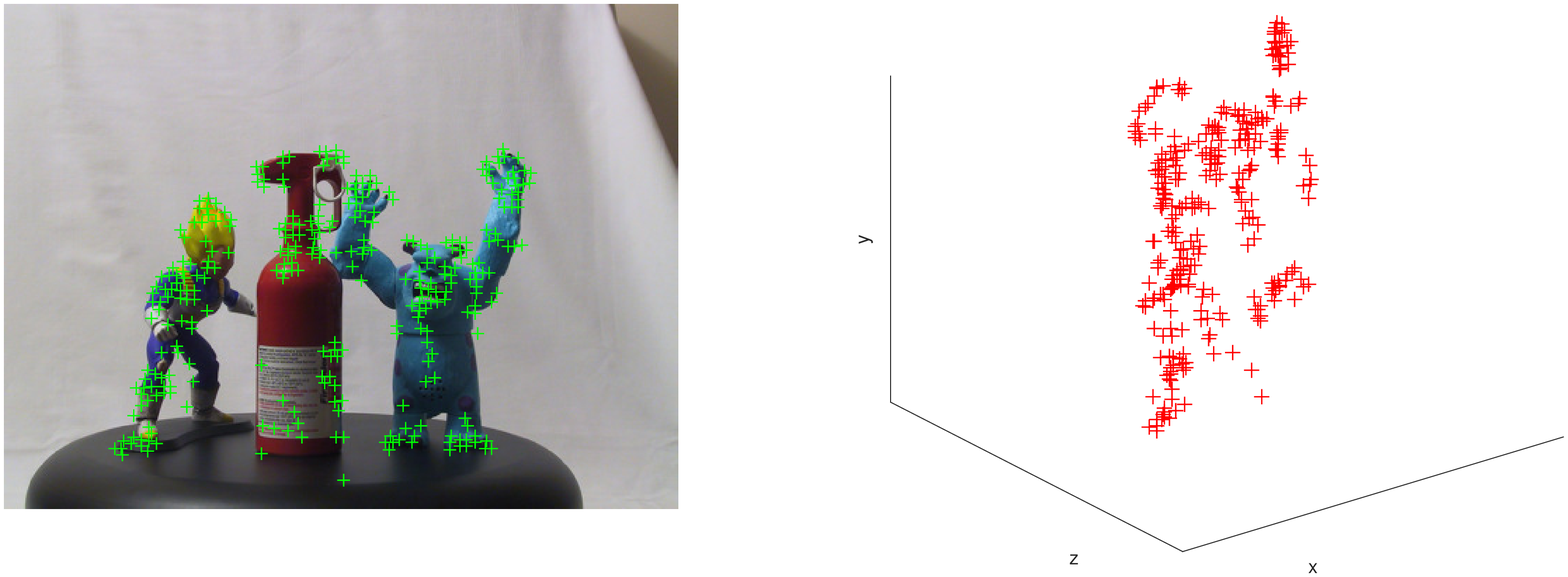}
		\caption{Standing (310 points)}
		\label{fig0d}
	\end{subfigure}
	\\
	\begin{subfigure}[h]{0.49\textwidth}
		\includegraphics[width=1\textwidth]{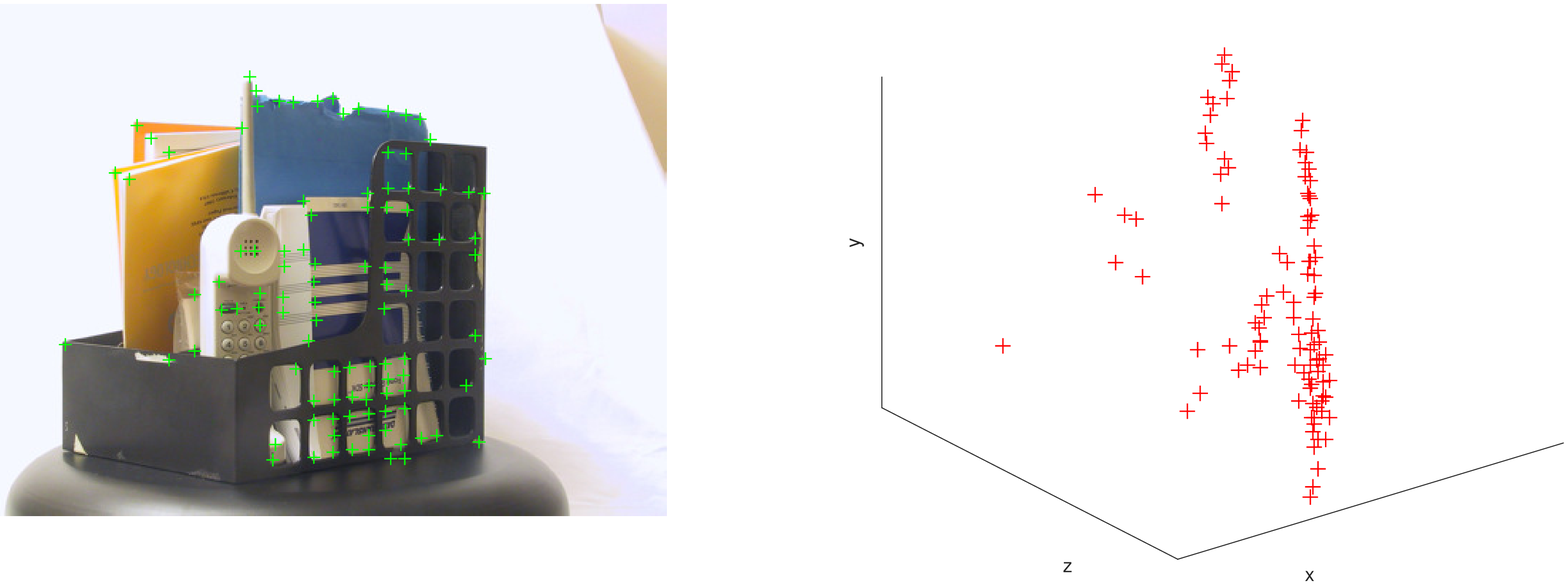}
		\caption{StorageBin (102 points)}
		\label{fig0e}
	\end{subfigure}
	\caption{The Caltech Turntable dataset objects used in~\cite{yoon2012} and the centralized SVD-based affine structure from motion result. Green dots on the image frame show the feature points tracked. All objects were tracked for 30 frames. The frames are distributed evenly to the 5 cameras.}
	\label{fig:caltech0}
\end{figure*}

We present example image frames from the Caltech Turntable~\cite{Moreels2007} dataset used in~\cite{yoon2012}. We compare the proposed methods, ADMM with Varying Penalty (ADMM-VP), ADMM with Adaptive Penalty (ADMM-AP) and ADMM with Network Adaptive Penalty (ADMM-NAP) and their combination (ADMM-VP + AP, ADMM-VP + NAP) with the standard ADMM based D-PPCA~\cite{yoon2012} using the same experimental setting. Fig.~\ref{fig:caltech0} shows an example frame, feature points extracted from the frame and the centralized SVD-based reconstructed structure we used as ground truth. In the paper, we showed the results of Standing. 

Fig.~\ref{fig:caltech} summarizes the results on the remaining four objects. The findings and analysis explained in the main paper on the object Standing also apply to these four remaining objects. First, we compare the top and the middle rows. One can see that when the graph is less connected (ring, top row) the proposed adaptive penalty method can boost ADMM-VP which cannot utilize the full residual information of fully connected case (complete, middle row), as explained in synthetic data experiments. 

Second, we compare the middle and the bottom rows. The network topologies are the same as complete but $t^{\max}$ value required for ADMM-VP, ADMM-AP, ADMM-VP + AP is different from these two groups of experiments. When $t^{max} = 50$ (middle row), all methods can accelerate throughout the iterations. However, when $t^{max} = 5$ (bottom row), the methods that depend on $t^{max}$ cannot accelerate after 5 iterations thus show similar behaviour as the baseline ADMM. On the other hand, ADMM-NAP based methods could accelerate by adaptively modifying the maximum number of penalty updates.

%
%

\begin{figure*}[t]
\centering
$\begin{array}{r l c c c c}
	\small \rotatebox[origin=c]{90}{Ring} \normalsize
	&
	\small \rotatebox[origin=c]{90}{$t^{max} = 50$} \normalsize
	&
	\begin{subfigure}[h]{0.2\textwidth}
		\includegraphics[width=1\textwidth]{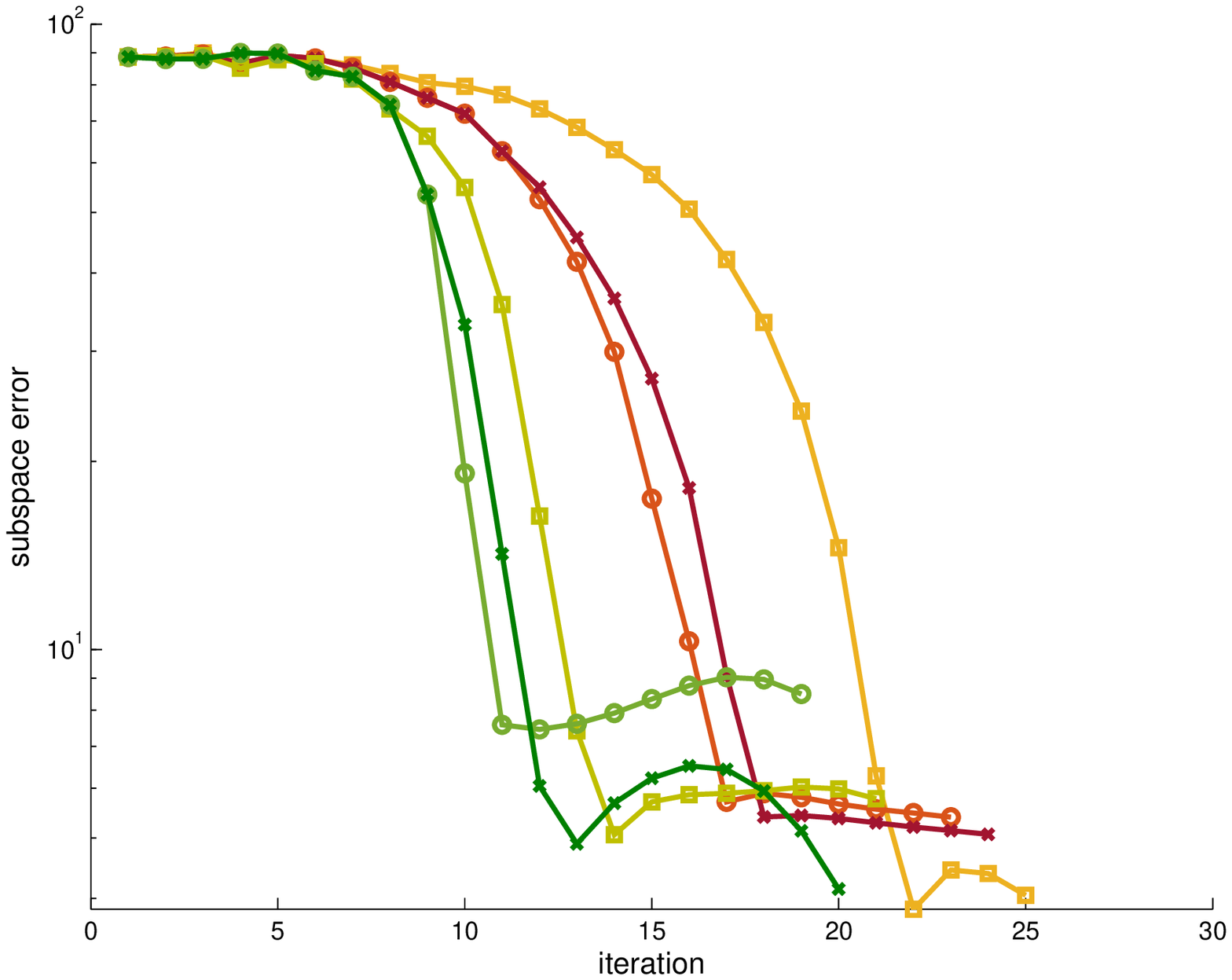}
		\label{fig41}
	\end{subfigure}
	&
	\begin{subfigure}[h]{0.2\textwidth}
		\includegraphics[width=1\textwidth]{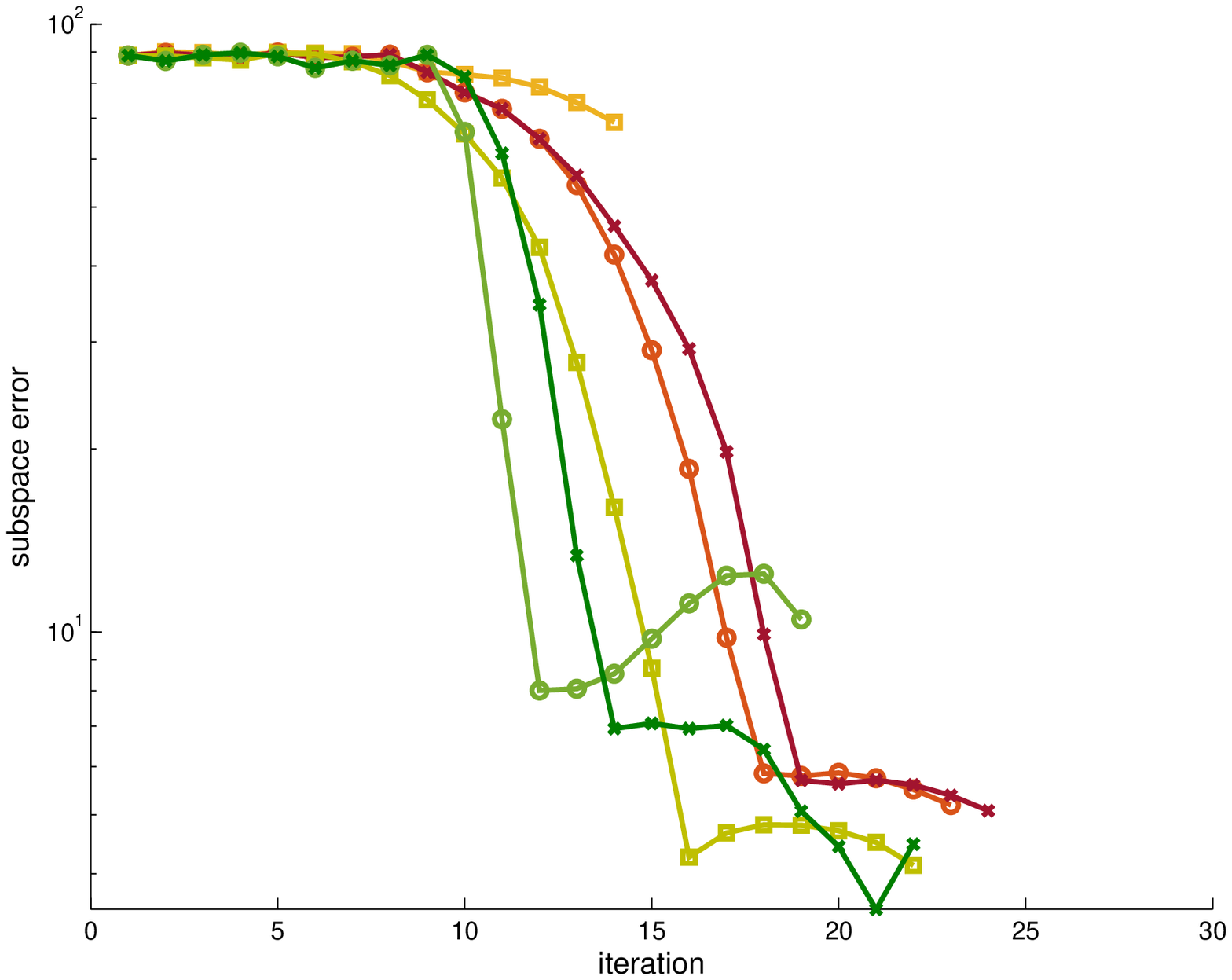}
		\label{fig42}
	\end{subfigure}
	&
	\begin{subfigure}[h]{0.2\textwidth}
		\includegraphics[width=1\textwidth]{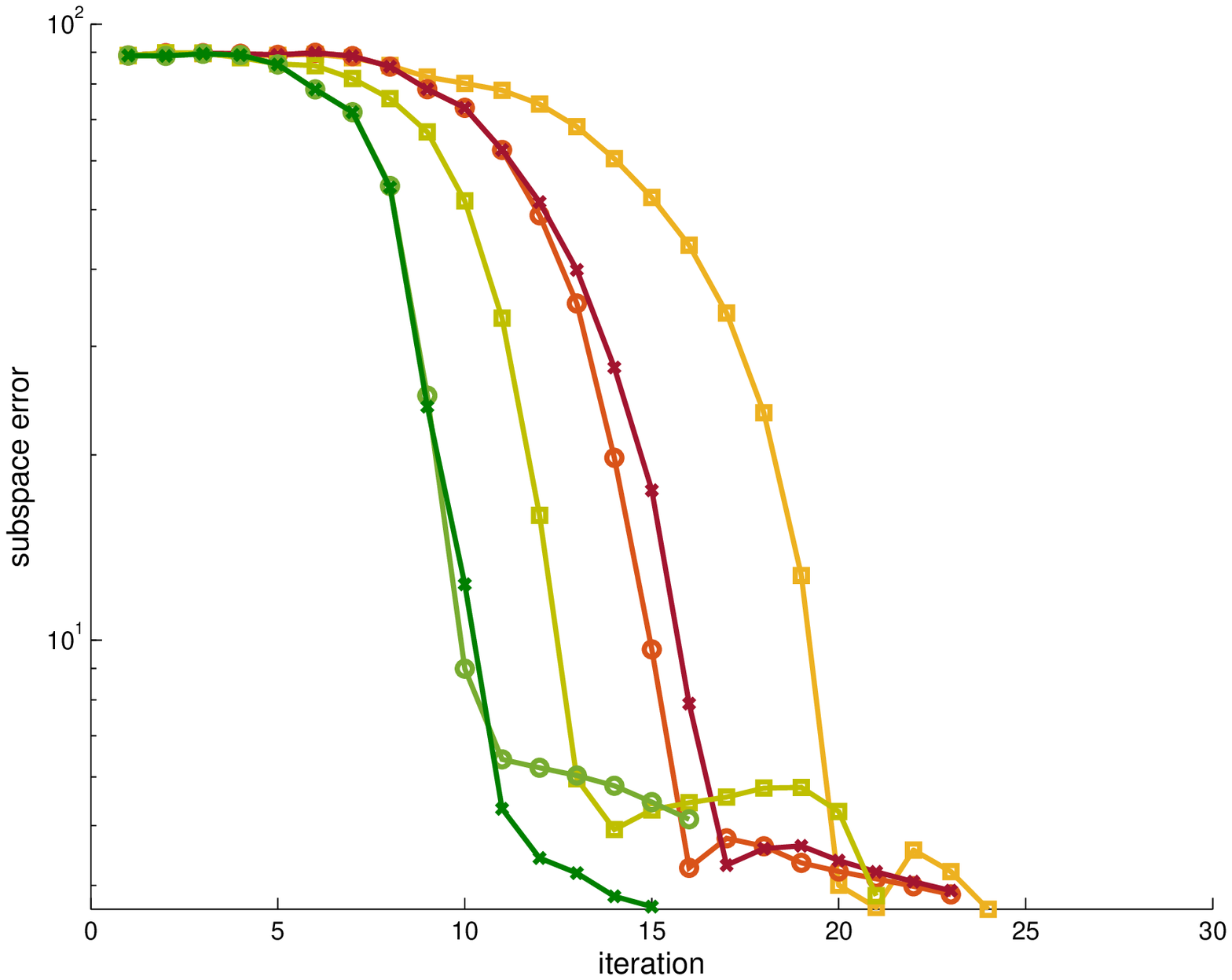}
		\label{fig43}
	\end{subfigure}
	&
	\begin{subfigure}[h]{0.2\textwidth}
		\includegraphics[width=1\textwidth]{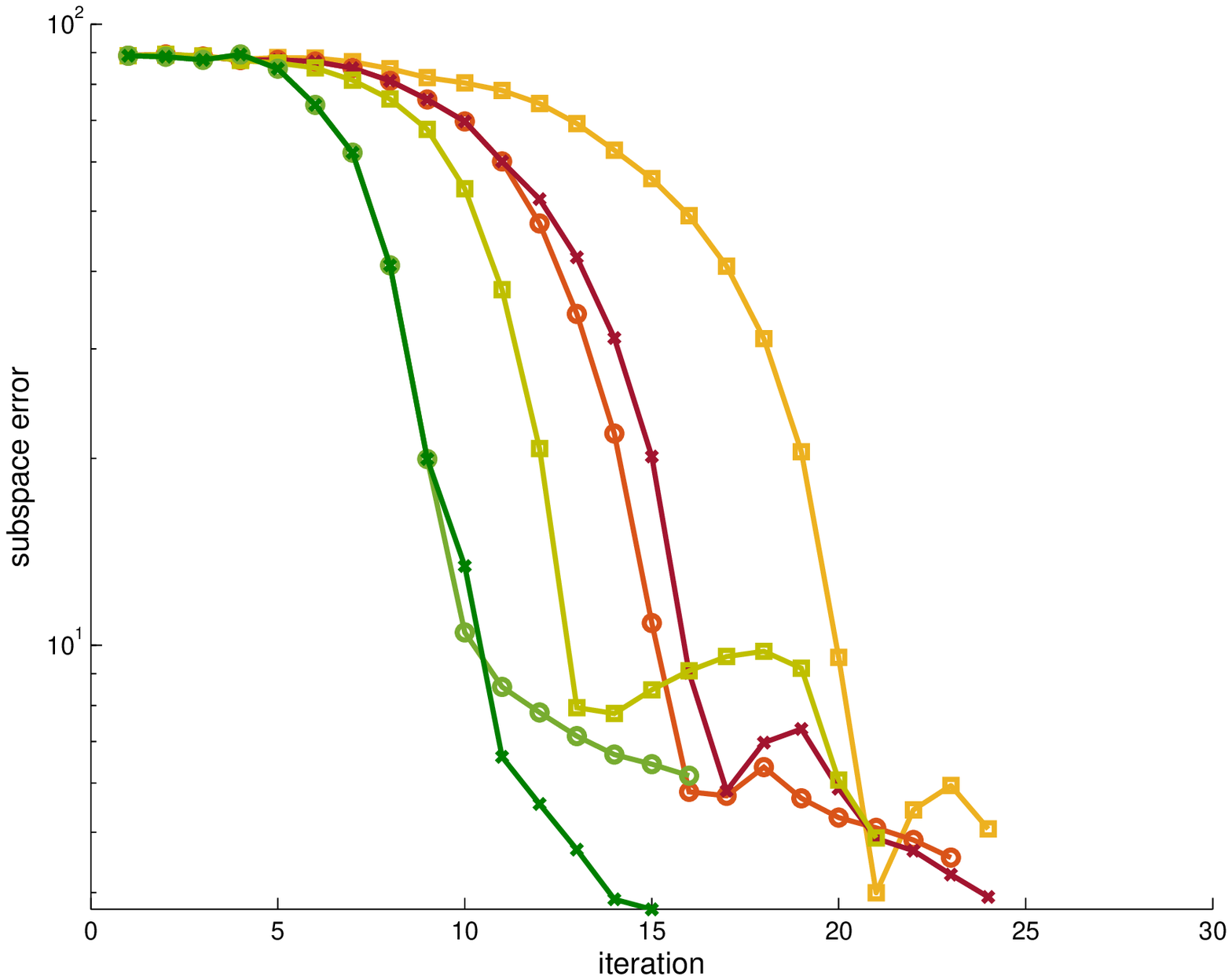}
		\label{fig45}
	\end{subfigure}
	\\
	\rotatebox[origin=c]{90}{Complete}
	&
	\rotatebox[origin=c]{90}{$t^{max} = 50$}
	&
	\begin{subfigure}[h]{0.2\textwidth}
		\includegraphics[width=1\textwidth]{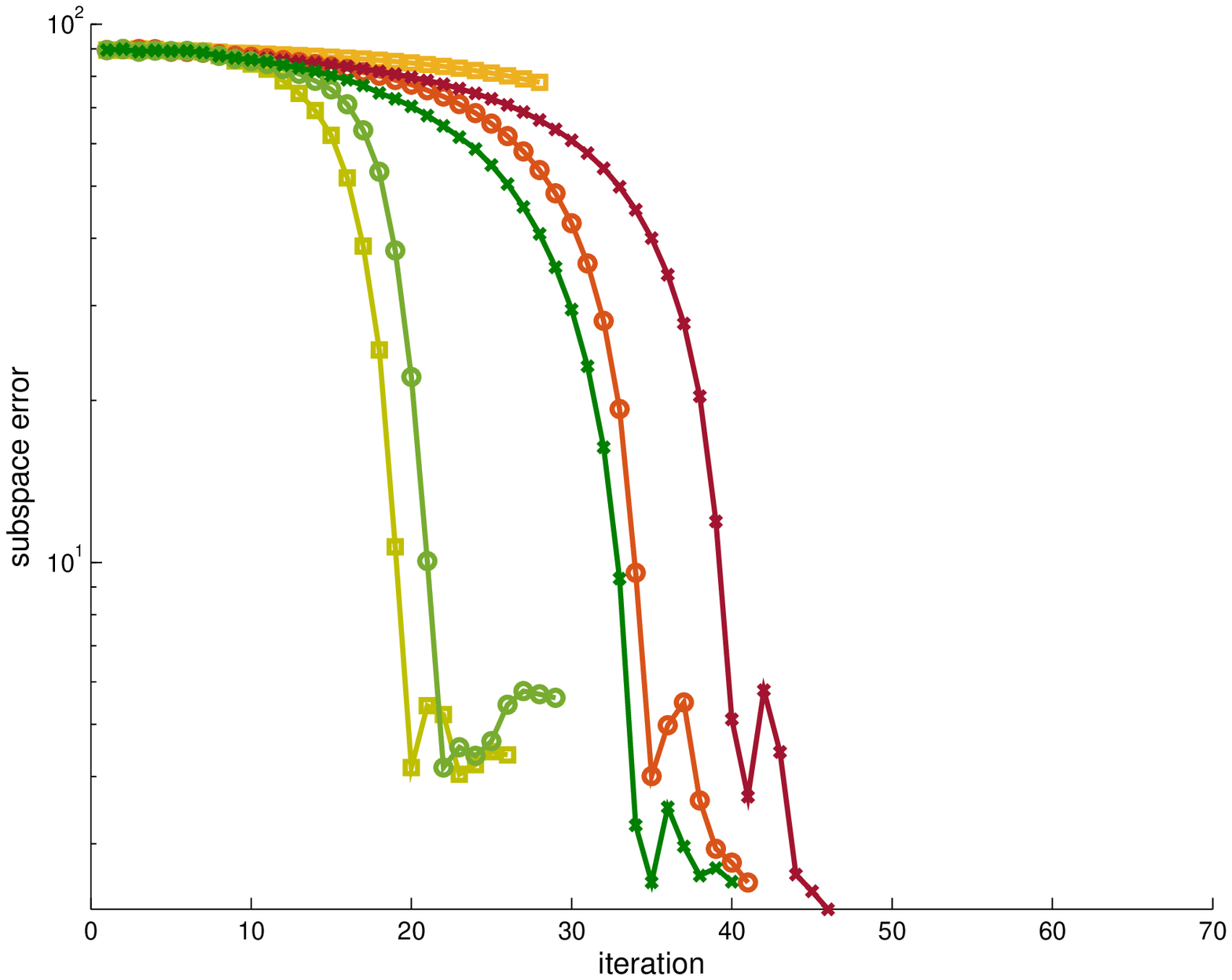}
		\label{fig41}
	\end{subfigure}
	&
	\begin{subfigure}[h]{0.2\textwidth}
		\includegraphics[width=1\textwidth]{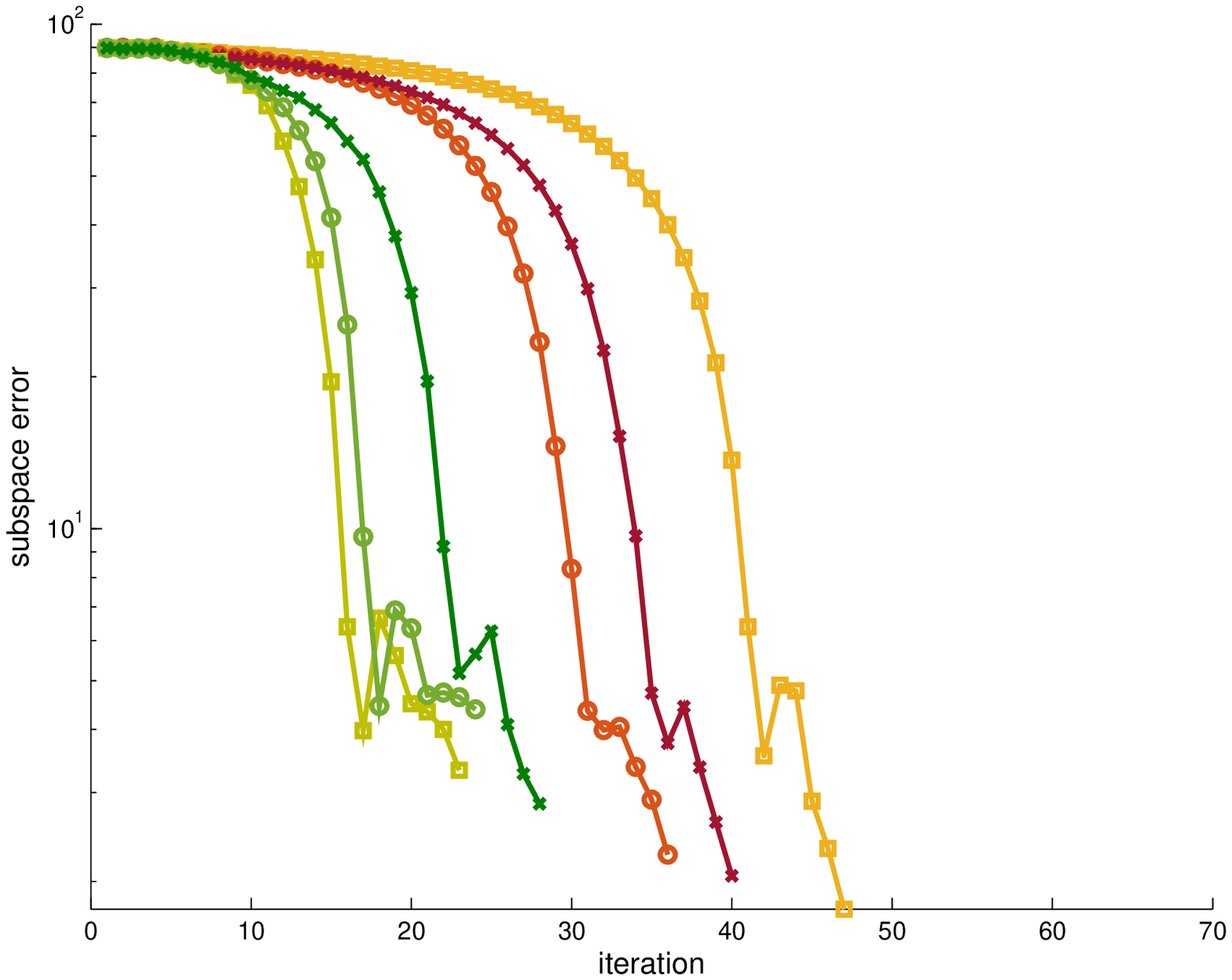}
		\label{fig42}
	\end{subfigure}
	&
	\begin{subfigure}[h]{0.2\textwidth}
		\includegraphics[width=1\textwidth]{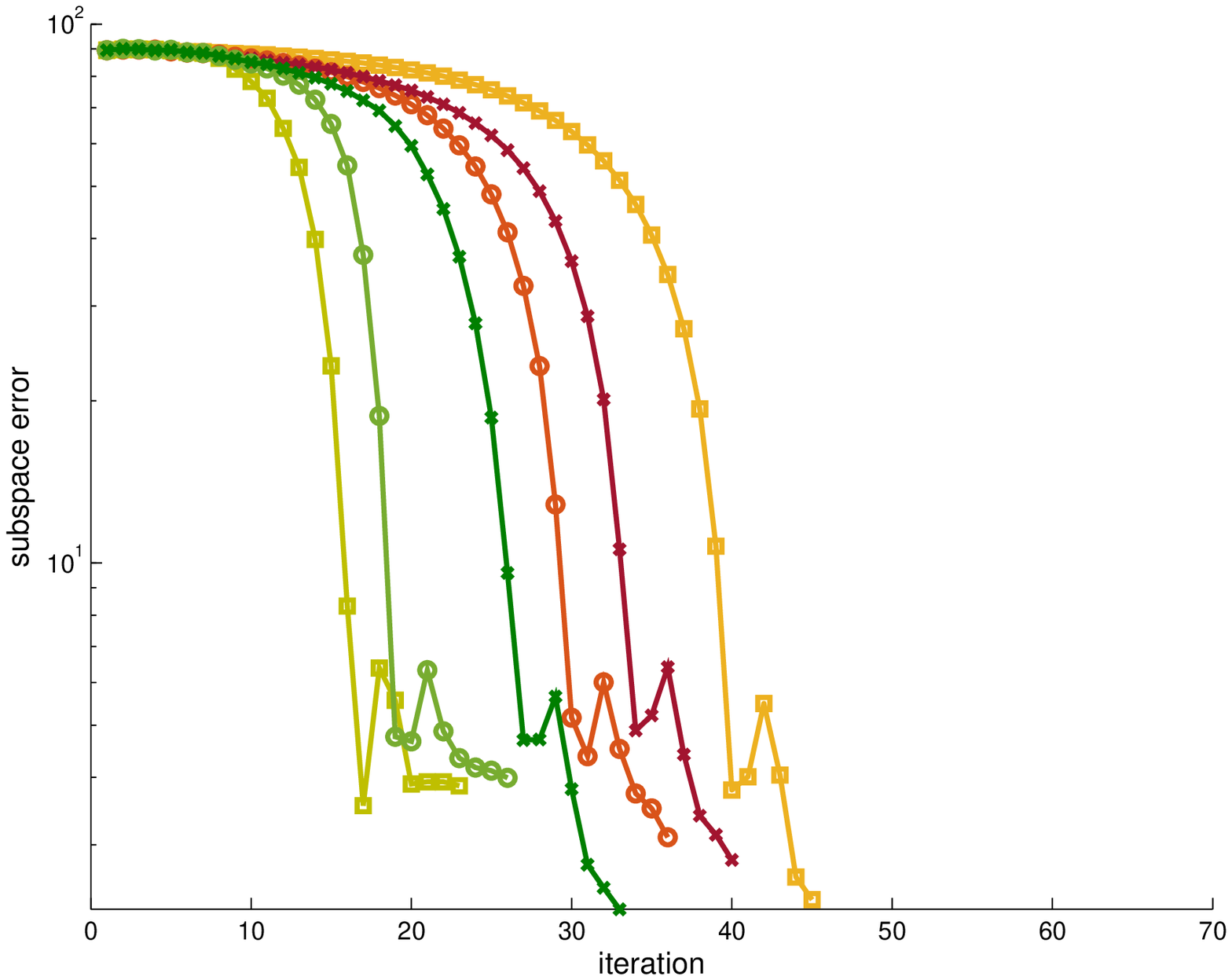}
		\label{fig43}
	\end{subfigure}
	&
	\begin{subfigure}[h]{0.22\textwidth}
		\includegraphics[width=1\textwidth]{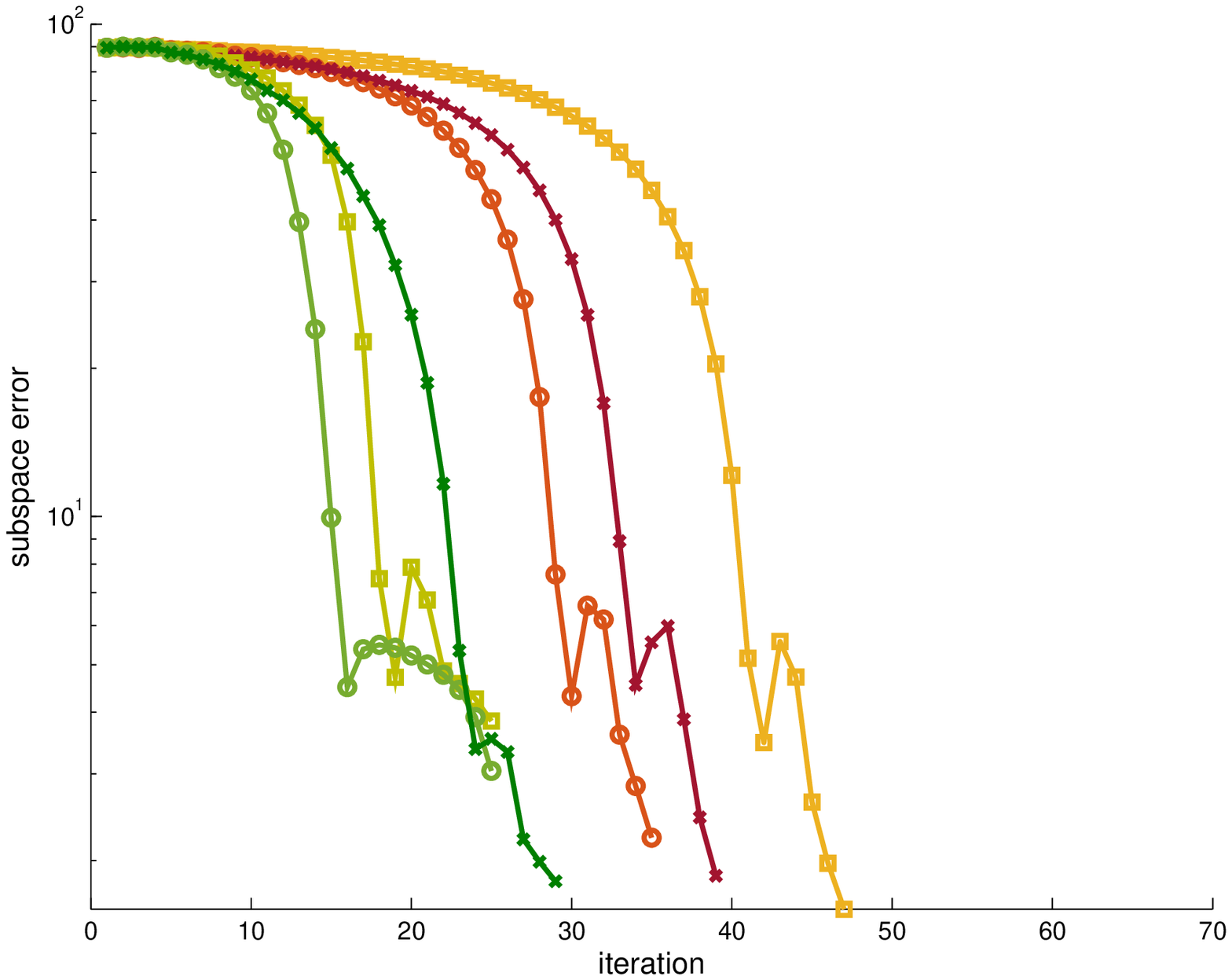}
		\label{fig45}
	\end{subfigure}
	\\
	\rotatebox[origin=c]{90}{Complete}
	&
	\rotatebox[origin=c]{90}{$t^{max} = 5$}
	&
	\begin{subfigure}[h]{0.2\textwidth}
		\includegraphics[width=1\textwidth]{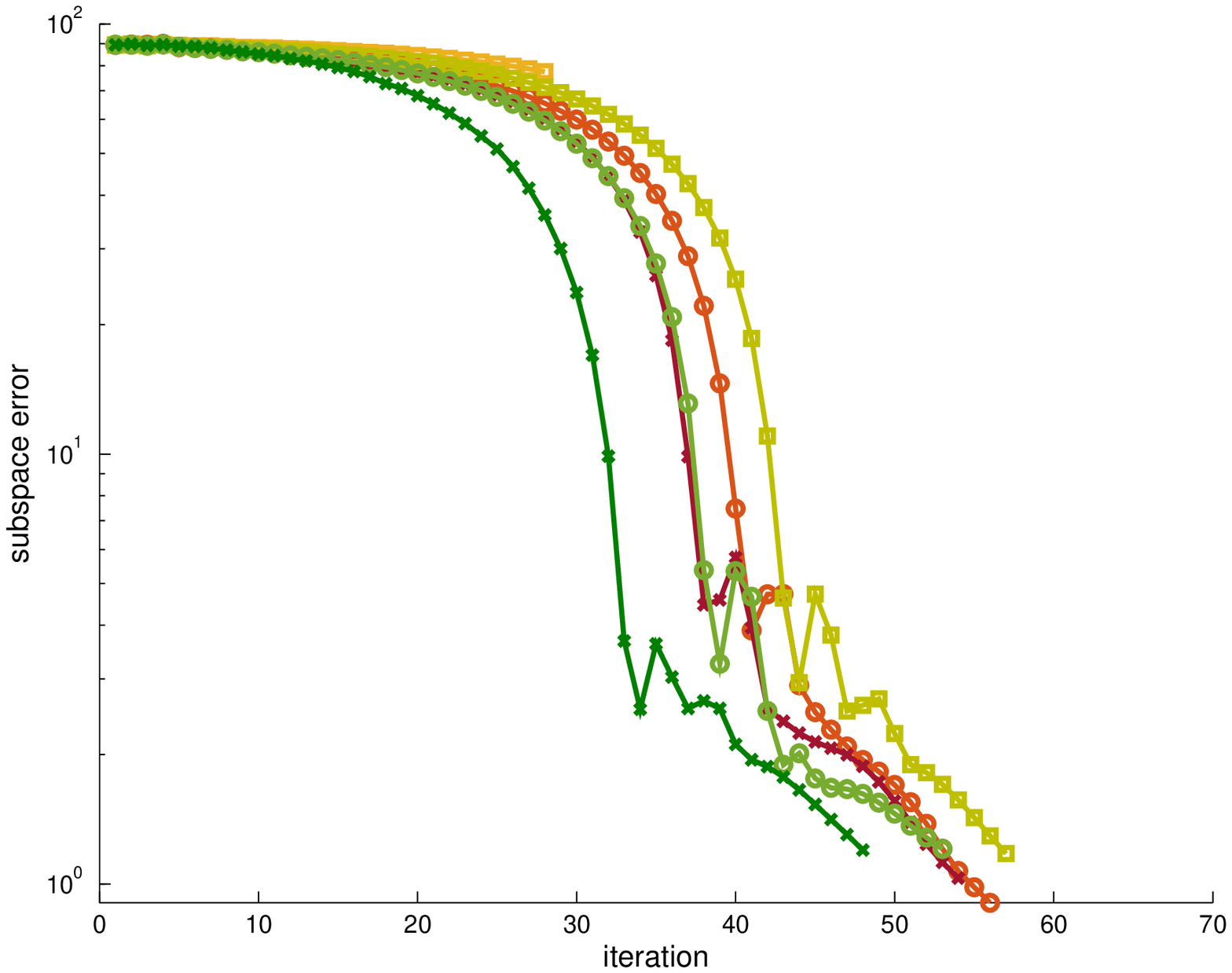}
		\label{fig41}
	\end{subfigure}
	&
	\begin{subfigure}[h]{0.2\textwidth}
		\includegraphics[width=1\textwidth]{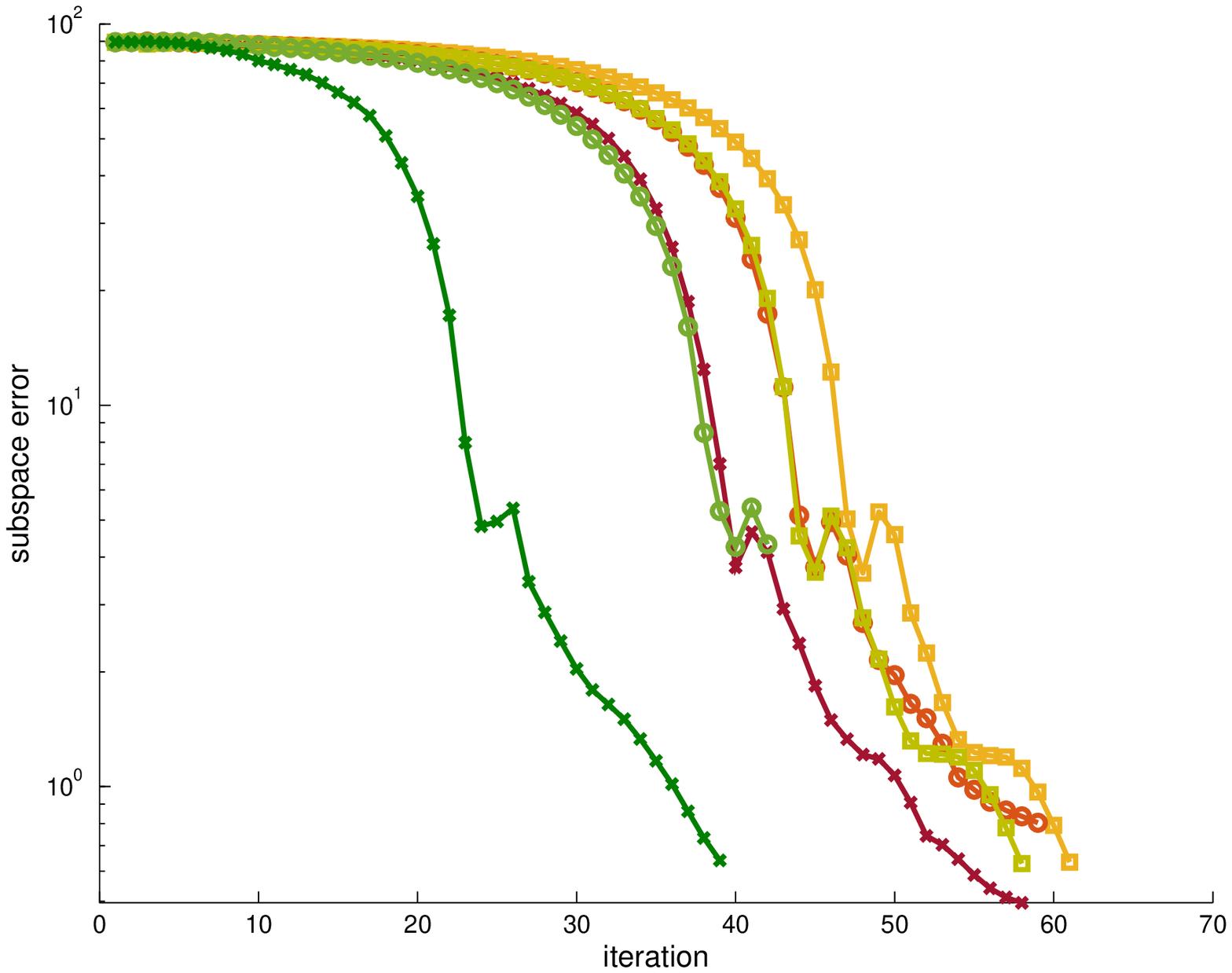}
		\label{fig42}
	\end{subfigure}
	&
	\begin{subfigure}[h]{0.2\textwidth}
		\includegraphics[width=1\textwidth]{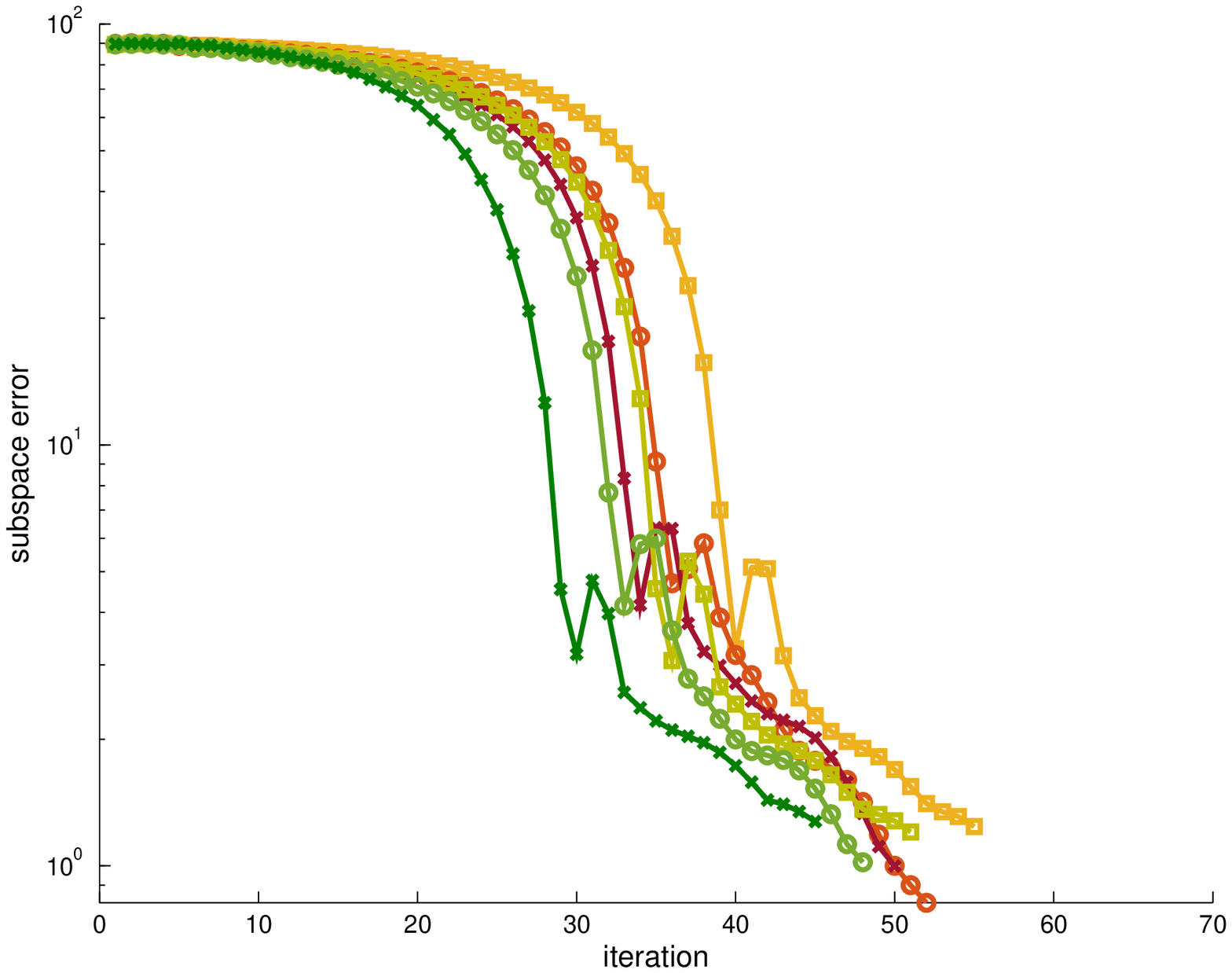}
		\label{fig43}
	\end{subfigure}
	&
	\begin{subfigure}[h]{0.2\textwidth}
		\includegraphics[width=1\textwidth]{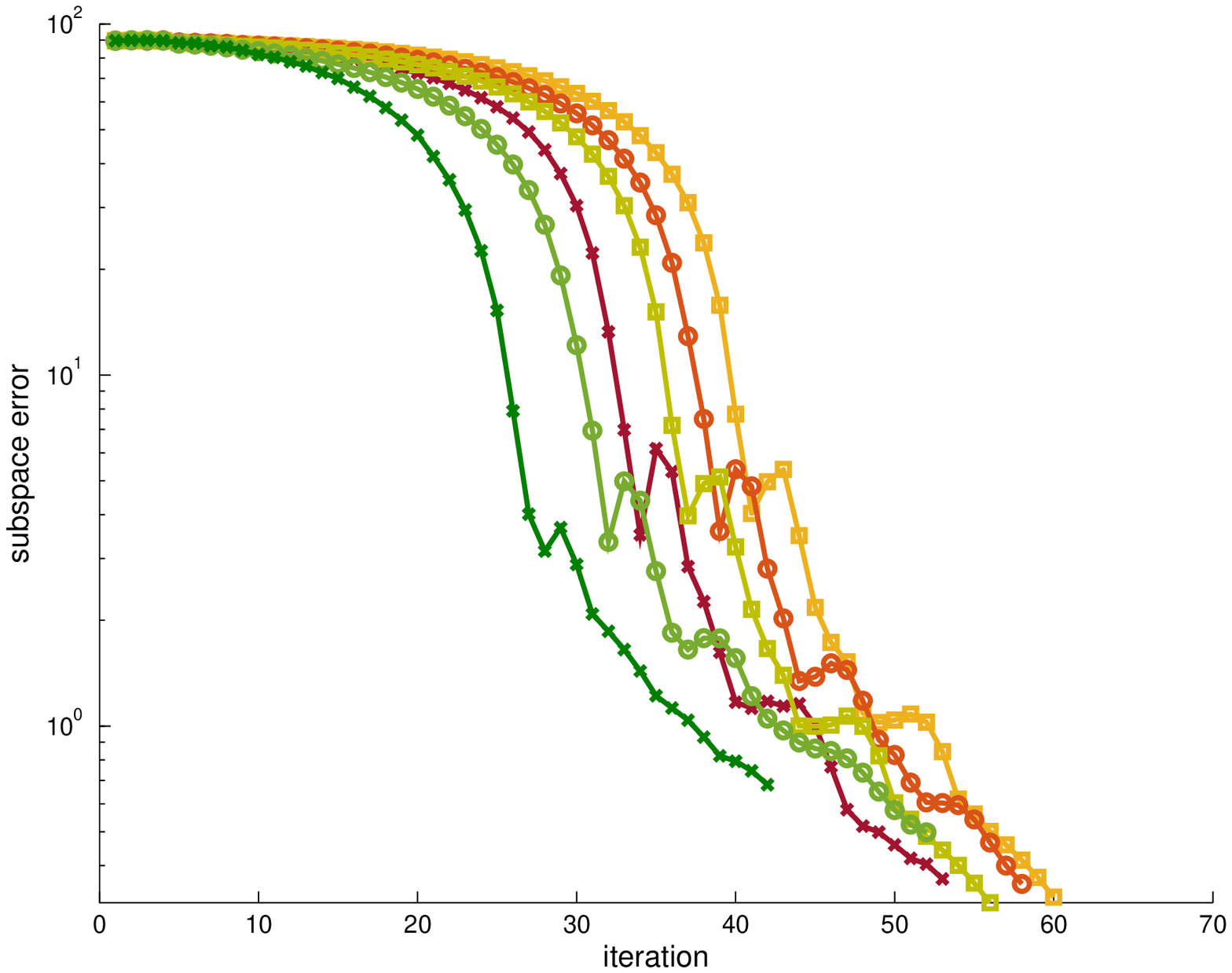}
		\label{fig45}
	\end{subfigure}
    \\
	& &
	\text{\small{(a) BallSander}} &
	\text{\small{(b) BoxStuff}}   &
	\text{\small{(c) Rooster}}    &
	\text{\small{(d) StorageBin}}  
	\\
	& &
	~~~~~\text{\small{($62$ points)}} & 
	~~~~~~\text{\small{($67$ points)}} & 
	~~~~~~~\text{\small{($189$ points)}} & 
	~~~~~\text{\small{($102$ points)}}
\end{array}$
	\caption{The comparison of proposed methods and the baseline ADMM using the subspace angle error of the reconstructed 3D structure with different objects in Caltech dataset. (top) $t^{max} = 50$, ring, (middle) $t^{max} = 50$, complete, (bottom) $t^{\max} = 5$, complete network. Refer Fig. 2 in the main paper for the labels.}
	\label{fig:caltech}
\end{figure*}

\clearpage
\small{}

\bibliographystyle{unsrt}
\bibliography{reference}

\end{document}